\documentclass[final,5p,times,twocolumn,authoryear]{elsarticle}

\usepackage{amsmath,amssymb,amsfonts}
\usepackage{graphicx}
\usepackage{adjustbox}
\usepackage{colortbl}
\usepackage{multirow}
\usepackage{textcomp}
\usepackage{booktabs}
\usepackage{algorithm}
\usepackage{tikz}
\usetikzlibrary{shapes.geometric, arrows.meta, positioning, calc, fit}
\usepackage{pgfplots}
\pgfplotsset{compat=1.18}
\usepackage[T1]{fontenc}
\usepackage{newtxtext,newtxmath}
\usepackage{comment}
\usepackage{caption}
\usepackage{algpseudocode}
\usepackage{float} 
\usepackage[utf8]{inputenc}
\usepackage{pifont} 
\PassOptionsToPackage{dvipsnames}{xcolor}
\usepackage{xcolor}
\usepackage{listings}
\UseRawInputEncoding
\usepackage{bm} 
\usepackage{adjustbox}

\definecolor{commentgreen}{rgb}{0,0.6,0}
\definecolor{green}{rgb}{0,0.50,0.10}

\newcommand{\revision}[1]{#1}
\newenvironment{revisionblock}{}{}

\usepackage[fontsize=9pt]{fontsize}
\usepackage{fancyhdr}
\fancypagestyle{firstpage}{\fancyhf{}}

\newcommand{\tabfont}{\fontsize{7.4}{8.9}\selectfont}
\newcommand{\fitcol}[1]{\adjustbox{max width=\columnwidth}{#1}}
\newcommand{\fitpage}[1]{\adjustbox{max width=\textwidth}{#1}}

\usepackage[colorlinks=true,linkcolor=blue,citecolor=blue,urlcolor=blue]{hyperref}

\begin{document}

\thispagestyle{firstpage}

\newcommand{\farclussfrontblock}{%
\noindent{\fontsize{9.5}{12}\selectfont Full Length Article\par}
\vspace{12pt}

\noindent{\fontsize{17}{20}\selectfont\raggedright\hyphenpenalty=10000\exhyphenpenalty=10000\relax
 FARCLUSS: Fuzzy Adaptive Rebalancing and
Contrastive Uncertainty Learning for Semi-Supervised Semantic Segmentation\par}
\vspace{11pt}

\noindent{\fontsize{12}{16}\selectfont Ebenezer Tarubinga\,\textsuperscript{a},
Jenifer Kalafatovich\,\textsuperscript{a},
Seong-Whan Lee\,\textsuperscript{a,}\textsuperscript{$\ast$}\par}
\vspace{8pt}

\noindent{\fontsize{7.6}{10}\selectfont\itshape
\textsuperscript{\upshape a}\,Department of Artificial Intelligence,
Korea University, Seoul, Korea\par}
\vspace{11pt}

\noindent\rule{\textwidth}{0.5pt}
\vspace{9pt}

\noindent\begin{minipage}[t]{0.257\textwidth}
  \fontsize{8}{10.5}\selectfont
  A\,R\,T\,I\,C\,L\,E\ \ I\,N\,F\,O\par
  \vspace{3pt}\noindent\rule{\linewidth}{0.4pt}\par
  \vspace{5pt}
  \textit{Keywords:}\par
  Semi-supervised learning\par
  Semantic segmentation\par
  Fuzzy pseudo-labeling\par
  Contrastive learning\par
  Uncertainty estimation\par
\end{minipage}%
\hspace{0.057\textwidth}%
\begin{minipage}[t]{0.681\textwidth}
  \fontsize{8}{10.5}\selectfont
  A\,B\,S\,T\,R\,A\,C\,T\par
  \vspace{3pt}\noindent\rule{\linewidth}{0.4pt}\par
  \vspace{5pt}
  \fontsize{8.4}{11}\selectfont
  Semi-supervised semantic segmentation (SSSS) faces persistent challenges in effectively leveraging unlabeled data, including ineffective utilization of pseudo-labels, the exacerbation of class imbalance biases, and the neglect of prediction uncertainty. Moreover, current approaches often discard uncertain regions through strict thresholding, thereby favouring dominant classes. To address these limitations, we introduce a holistic framework that transforms uncertainty into a learning asset through four principal components: (1) fuzzy pseudo-labeling, which preserves soft class distributions from top-$K$ predictions to enrich supervision; (2) uncertainty-aware dynamic weighting, that modulates pixel-wise contributions via entropy-based reliability scores; (3) adaptive class rebalancing, which dynamically adjusts the loss to counteract long-tailed class distributions; and (4) lightweight contrastive regularization, that encourages compact and discriminative feature embeddings. Extensive experiments on Pascal VOC and Cityscapes demonstrate that our method outperforms current state-of-the-art approaches, with pronounced gains on under-represented classes and ambiguous boundary regions. Our code is available at \url{https://github.com/psychofict/FARCLUSS}.
\end{minipage}

\vspace{11pt}
\noindent\rule{\textwidth}{0.5pt}
\vspace{4pt}
}

\twocolumn[\farclussfrontblock]

\begingroup
\renewcommand{\thefootnote}{}
\footnotetext{\hspace*{-1.6em}$^{\ast}$\,Corresponding author.}
\footnotetext{\raggedright\hspace*{-1.6em}\textit{E-mail addresses:}
  \texttt{psychofict@korea.ac.kr} (E. Tarubinga),
  \texttt{jenifer@korea.ac.kr} (J. Kalafatovich),
  \texttt{sw.lee@korea.ac.kr} (S.-W. Lee).}
\footnotetext{\raggedright\hspace*{-1.6em}Accepted for publication in \textit{Neural
  Networks} (Elsevier). This is the authors' accepted manuscript; it has not
  gone through Elsevier's proof correction and may differ from the published
  version of record.}
\endgroup
\addtocounter{footnote}{0}

\section{Introduction}

Semantic segmentation, defined as the process of assigning class labels to every pixel in an image, supports a range of critical applications, including autonomous driving and medical diagnostics. By facilitating pixel-level scene analysis, it supports systems that demand fine-grained environmental understanding \citep{tarubinga2025confidence}. Nonetheless, fully supervised segmentation techniques usually depend on extensive datasets with detailed annotations; making it costly in fields like urban scene analysis or healthcare, where manual labelling is both expensive and time-consuming \citep{10.1109/cvpr.2019.00262}. To mitigate this dependency on labeled data, semi-supervised semantic segmentation (SSSS) has emerged as a promising alternative. By leveraging a modest set of labeled images alongside abundant unlabeled data, SSSS frameworks strive to achieve performance comparable to fully supervised models while substantially reducing annotation requirements. Early work in this area centered on pseudo-labeling \citep{lee2013pseudo}, wherein a teacher model generates approximate labels for unlabeled data that a student model subsequently learns from. Later research introduced consistency regularization \citep{laine2017temporal}, enforcing prediction invariance under various geometric or photometric transformations. Although these techniques have shown improvements, they remain constrained by three principal issues: suboptimal use of pseudo-labels, class imbalance in long-tailed datasets, and inadequate handling of prediction uncertainty.

Although recent approaches attempt to address these issues, they often lack comprehensive solutions. Most methods treat pseudo-label uncertainty as a liability. They discard low-confidence predictions via strict thresholding, overlooking subtle probabilistic relationships among plausible classes. For instance, a pixel at the boundary of `sidewalk' and `road' may carry near-equal probabilities for both classes, yet is often removed, forfeiting valuable supervisory signals. UniMatch \citep{yang2023revisiting} combines feature- and output-level consistency but relies on fixed confidence thresholds to generate hard pseudo-labels. As a result, ambiguous predictions are eliminated, especially near object boundaries or occluded regions, leading to loss of information.

Class imbalance remains inadequately addressed. Although some methods adjust global thresholds, they fail to rebalance losses based on per-batch class distributions, perpetuating bias toward majority classes. PS-MT \citep{liu2022perturbed} employs progressive self-training with momentum teachers and adaptive thresholds but lacks explicit mechanisms for class rebalancing. Consequently, dominant categories (e.g., ``road'' or ``sky'' in urban scenes) are over-represented in pseudo-labels.

Methods that use contrastive learning to address previous mentioned limitations often incur high computational cost due to extensive feature alignments, limiting scalability on large datasets. ReCo \citep{liu2022reco} improves feature discriminability using region-level contrastive learning, but its pairwise embedding comparisons introduce considerable overhead. Cross-pseudo supervision (CPS) \citep{chen2021semisupervised} trains dual networks that supervise each other, yet applies uniform weighting to pseudo-labels, neglecting the need to treat uncertain regions differently.

This work proposes a unified framework that systematically addresses common challenges in semi-supervised segmentation, including pseudo-label noise, class imbalance, and computational overhead. First, we introduce fuzzy pseudo-labeling, which derives soft label distributions from the top-$K$ class probabilities at each pixel. By preserving uncertainty (unlike previous methods' strict thresholding), our method treats ambiguity as a constructive supervisory cue. This enables the model to learn from regions like unclear boundaries (e.g., pavement edges between ``car'' and ``road'') instead of discarding them.

Second, we implement uncertainty-aware dynamic weighting, assigning pixel-wise weights based on normalized entropy to modulate pseudo-label influence during training. This mechanism curbs the impact of noisy high-entropy regions (e.g., overlapping objects) while amplifying the contribution of confident predictions. This mechanism improves upon CPS by replacing uniform weighting with a more discriminative strategy \citep{chen2021semisupervised}.

Third, we propose adaptive class rebalancing to mitigate class imbalance by dynamically adjusting loss weights according to batch-level pseudo-label frequencies. This ensures under-represented categories like ``traffic sign'' receive adequate optimization focus without manual tuning.

Finally, we introduce lightweight contrastive regularization based on a prototype-centric design. By computing class-specific centroids from confidently labeled pixels, our approach promotes discriminative feature clustering while avoiding the computational burden of exhaustive pairwise comparisons as in ReCo \citep{liu2022reco}.

While existing methods have tackled individual challenges such as ineffective pseudo-label utilization, class imbalance, and high computational cost, they often fall short of addressing these issues simultaneously within a unified framework. The main contributions of this work are summarized as follows:

\vspace{1mm}
\begin{itemize}
\item \textbf{Fuzzy Pseudo-Labeling:} Constructs soft label distributions from top-$K$ class probabilities, preserving ambiguity as a learning signal rather than discarding uncertain pixels.
\item \textbf{Uncertainty-Aware Dynamic Weighting:} Modulates pixel-wise pseudo-label impact using normalized entropy, reducing noise from ambiguous regions and enhancing learning from confident predictions.
\item \textbf{Adaptive Rebalancing:} Dynamically scales losses based on pseudo-label frequencies per batch, improving learning for minority classes without manual heuristics.
\item \textbf{Lightweight Contrastive Regularization:} Prototype-based contrastive learning via class centroids avoids costly pairwise operations while promoting feature compactness.
\end{itemize}

\section{Related Works}

\subsection{Teacher-Student Frameworks and Pseudo-Label Refinement}
Semi-supervised semantic segmentation (SSSS) has progressed remarkably through teacher-student paradigms, wherein models generate pseudo-labels from unlabeled data to guide training. Early work such as FixMatch \citep{sohn2020fixmatch} employed confidence thresholding to filter pseudo-labels, leading to high-confidence yet inaccurate predictions reinforced model errors (confirmation bias). Cross Pseudo Supervision (CPS) \citep{chen2021semisupervised} mitigated this issue by training two networks jointly in a supervised manner using each other's pseudo-labels, thereby diversifying the learning signal. PseudoSeg \citep{zou2020pseudoseg} enhanced reliability through augmented prediction fusion, while Directional Context-Aware Consistency (DC-Loss) \citep{lai2021semi} encouraged alignment of class-specific features in differing contexts. More recent approaches, including the Error Localization Network (ELN) \citep{kwon2022semi} and $U^2\mathrm{PL}$ \citep{U2PL}, introduced modules to detect, filter, or reinterpret uncertain predictions as negative supervision. However, most of these methods still rely on strict thresholding, discarding ambiguous pixels and thereby limiting data usage. By contrast, our framework employs fuzzy pseudo-labeling to retain the top-K class probabilities for each pixel, preserving nuanced inter-class relationships and providing richer supervision even in low-confidence regions.

\subsection{Uncertainty-Aware Confidence Calibration}
Uncertainty estimation has become central to improving pseudo-label quality. For example, RePU \citep{zheng2021repunet} refined pseudo-labels using entropy-based uncertainty scores, and Bayesian Uncertainty Calibration \citep{cui2021bayesian} leveraged Monte Carlo dropout to identify unreliable predictions. Despite their effectiveness, these methods generally adopt uncertainty as a binary filter rather than as a learning signal. By contrast, our approach incorporates uncertainty-aware dynamic weighting, wherein normalized entropy modulates the contribution of each pseudo-label. High-entropy areas, such as boundaries or overlapping objects, are downweighted proportionately to their uncertainty, reducing noise while preserving partial supervisory information, a notable improvement over deterministic thresholding or uniform weighting.

\begin{figure*}[t]
\centering
\includegraphics[width=0.76\textwidth]{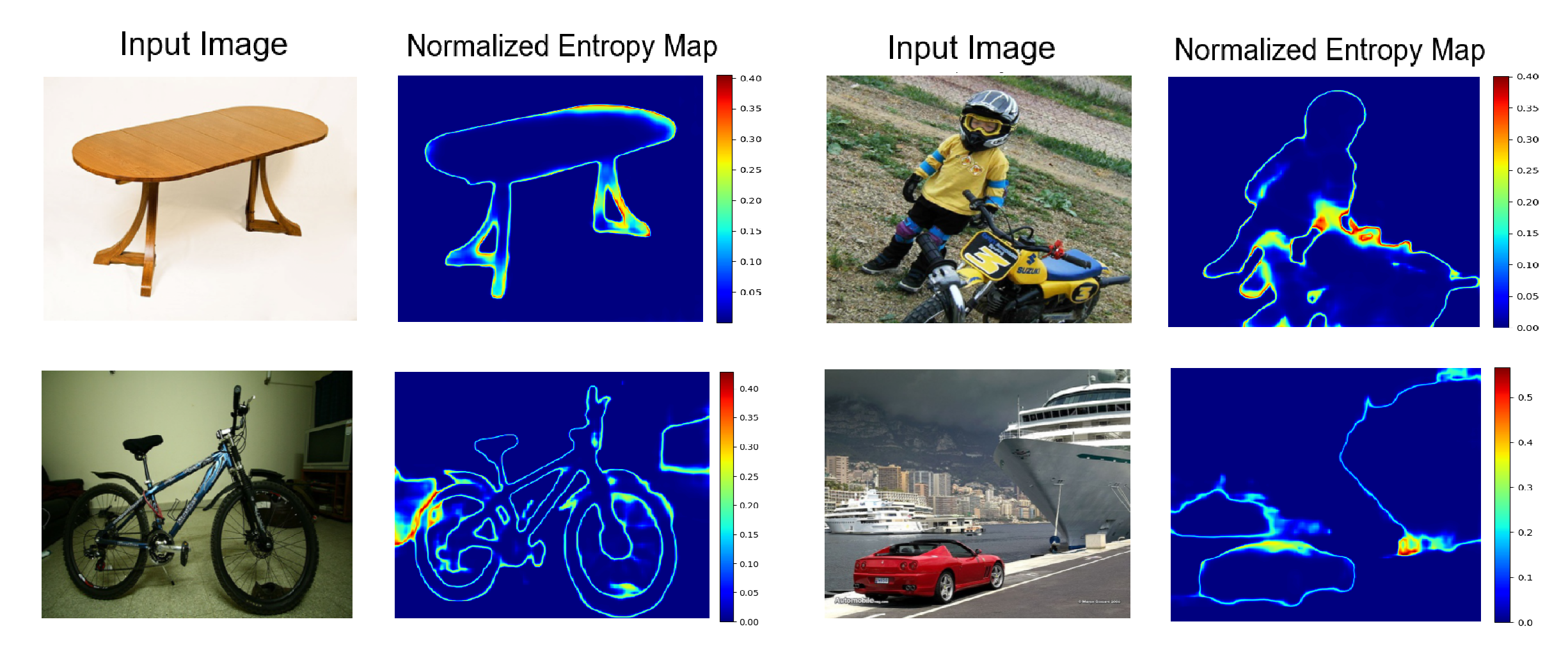}
\caption{Pixel-wise normalized entropy map from the teacher model on PASCAL VOC samples. Brighter colors represent higher prediction uncertainty. Uncertainty is concentrated at object boundaries and cluttered background regions.}
\label{entropy map.png}
\end{figure*}

\subsection{Class Imbalance and Rare Class Enhancement}
Class imbalance continues to pose significant challenges for SSSS, as pseudo-labels often exhibit a bias toward frequently occurring classes. DARS \citep{cho2021dars} aligned pseudo-label distributions with those from labeled data, and Adaptive Equalization Learning (AEL) \citep{hu2021semi} amplified underrepresented classes through adaptive loss weighting. Augmentation-based approaches, including Dynamic Cross-Set Copy-Paste (DCSCP) \citep{fan2022ucc} and iMAS \citep{zhao2023instance}, artificially increased rare-class samples or dynamically adjusted supervision based on confidence. While these methods successfully address class imbalance, they typically rely on global class distributions, overlooking variations at the batch level. In contrast, our adaptive class rebalancing technique leverages per-batch pseudo-label frequencies to scale losses, ensuring that classes like traffic sign in Cityscapes receive increased focus when they appear infrequently.  

\subsection{Computational Efficiency and Lightweight Learning}
Contemporary SSSS approaches frequently carry a high computational burden due to multi-model architectures or complex regularization. Cross-Consistency Training (CCT) \citep{ouali2020semi} reduced this burden by training multiple decoder heads with a shared encoder, while UniMatch \citep{yang2023revisiting} illustrated that a single-network FixMatch \citep{sohn2020fixmatch} can achieve performance comparable to multi-model solutions. Multi-Granularity Distillation (MGD) \citep{qin2022multi} further improved efficiency via hierarchical knowledge distillation. Building on these principles, our framework avoids dual-network configurations like CPS \citep{chen2021semisupervised} and auxiliary modules such as ELN \citep{kwon2022semi}. Through single-pass fuzzy pseudo-labeling, entropy-based weighting, and prototype-based contrastive learning, wherein class centroids derived from high-confidence pseudo-labels enhance intra-class compactness, we preserve scalability on large datasets without sacrificing segmentation accuracy. Unlike ReCo \citep{liu2022reco}, which entails computationally expensive pairwise comparisons, this prototype-based approach offers improved feature separability for minority classes with minimal overhead. More recently, PixCon \citep{tarubinga2026pixcon} argued that once a foundation-model teacher renders confidence filtering close to saturated, the accuracy that remains lies in how the embedding space is organised by class, and structures it with a clean-positive pixel-contrastive memory bank. FARCLUSS addresses the complementary regime of a conventional backbone, where thresholding still discards usable signal, and keeps its contrastive term prototype-based so that no per-pixel bookkeeping is required.


\section{Method}

\begin{figure*}[t]
\centering
\includegraphics[width=0.80\textwidth]{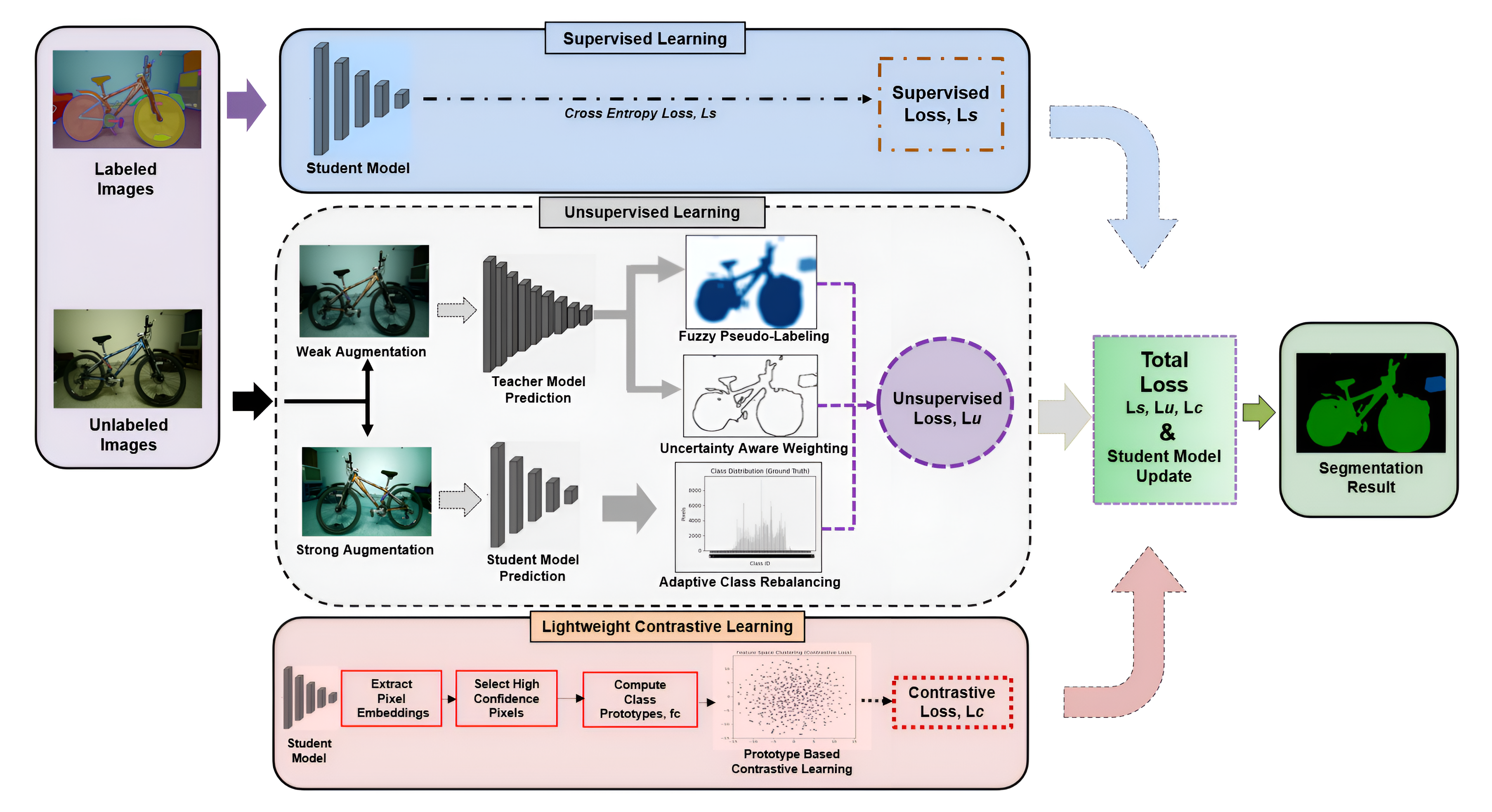}
\caption{Overview of the FARCLUSS Framework. FARCLUSS employs a teacher-student framework where labeled data supervise the student directly, while unlabeled data undergo fuzzy pseudo-labeling and uncertainty-based weighting via the teacher. Adaptive class rebalancing and prototype-based contrastive learning further refine the total loss.}
\label{fig1 framework.png}
\end{figure*}

Our method integrates fuzzy pseudo-labeling, uncertainty-based dynamic weighting, adaptive class rebalancing, and efficient contrastive regularization to address inefficient pseudo-label utilization, class imbalance biases, and underutilization of uncertain predictions. An overview of our proposed method is shown in Fig. \ref{fig1 framework.png} and the corresponding pseudocode is provided in Algorithm \ref{alg.1}.

\begin{algorithm*}[t]
  \caption{FARCLUSS pseudocode in PyTorch style}
\begin{lstlisting}[language=Python,linewidth=\textwidth,xleftmargin=0pt]
# f: encoder g + decoder h; f_ema: EMA teacher; aug_w/aug_s: perturbations
criterion_ce = nn.KLDivLoss(log_target=False, reduction="none")  
criterion_con = nn.CosineSimilarity(dim=1)
m = 0.99  # EMA momentum                                  # eq. 1
for x in loader_u:
    # generate weak/strong views
    xw = aug_w(x)
    xs1, xs2 = aug_s(xw), aug_s(xw)
    # teacher inference & fuzzy labels
    with torch.no_grad():
        pw = f_ema(xw)
        tk = pw.topk(K, 1).indices
        fuzzy = torch.zeros_like(pw).scatter_(1, tk, pw.gather(1, tk))
        fuzzy = fuzzy / fuzzy.sum(dim=1, keepdim=True) #eq. 2
        H = - (pw * pw.log()).sum(dim=1) / math.log(pw.shape[1])
        w_px = 1 - H  # pixel/instance weight              # eq. 6
    # student features & mask
    feat = g(torch.cat([xs1, xs2]))                 # B*2, C, H, W
    bs2, C, H, W = feat.shape
    m0 = torch.bernoulli(0.5 * torch.ones(bs2//2, C, device=feat.device))
    mask = torch.cat([m0, 1 - m0], 0).view(bs2, C, 1, 1)  
    pred = h(feat * mask)
    # supervised loss on labeled batch
    x_l, y_l = next(iter(loader_s))  
    loss_sup = nn.CrossEntropyLoss()(f(x_l).log_softmax(1), y_l)
    # adaptive class rebalancing loss
    lbl = fuzzy.argmax(1)
    cnt = lbl.bincount(minlength=pw.size(1)).float()
    w_cls = cnt.median() / (cnt + 1e-6) # eq. 7
    loss_u_map = criterion_ce(pred.log_softmax(1), fuzzy)  # B*2 x C x H x W
    loss_u = (w_px.view(-1,1,1,1) * w_cls[lbl].view(-1,1,1,1) * loss_u_map).mean()
    # prototype contrastive loss
    sel = w_px > 0.5
    fs = feat[sel]                                 # N_sel, C, H, W
    lb = lbl[sel]
    fs_vec = fs.mean((2,3))                             # N_sel x C
    cnt_sel = lb.bincount(minlength=pw.size(1)).float()
    proto = torch.zeros(pw.size(1), C, device=feat.device)
    proto.index_add_(0, lb, fs_vec)
    proto = proto / (cnt_sel.unsqueeze(1) + 1e-6)           # eq. 9
    loss_c = (1 - criterion_con(fs_vec, proto[lb])).mean() # eq. 10
    # update student & EMA
    loss = loss_sup + lam_u * loss_u + lam_c * loss_c        # eq. 11
    optimizer.zero_grad(); loss.backward(); optimizer.step()
    for pt, ps in zip(f_ema.parameters(), f.parameters()):
        pt.data.mul_(m).add_(ps.data, alpha=1-m)
\end{lstlisting}
  \label{alg.1}
\end{algorithm*}

\subsection{Teacher-Student Architecture}

We adopt a teacher-student framework inspired by classic mean-teacher approaches \citep{tarvainen2017mean}. Let $\mathcal{X}_l = \{(x_i^l, y_i^l)\}_{i=1}^{N_l}$ represent labeled training data, and $\mathcal{X}_u = \{x_j^u\}_{j=1}^{N_u}$ be unlabeled training data. The student model, parameterized by $\theta_s$, is trained with labeled data on $\mathcal{X}_l$ and with additional unlabeled data on $\mathcal{X}_u$. The teacher model, parameterized by $\theta_t$, provides pseudo-labels and is updated as an exponential moving average (EMA) of the student's weights. Specifically, after each training iteration $k$, weights are updated following:
\begin{equation}\label{eq:ema}
    \theta_t^{(k)} = \alpha \theta_t^{(k-1)} \;+\; (1 - \alpha)\,\theta_s^{(k)},
\end{equation}
where $\alpha=0.99$ is a momentum term that stabilizes the teacher's evolution. This design helps smooth out noisy student updates.

\subsection{Fuzzy Soft Pseudo-Labeling}

Traditional pseudo-labeling strategies often rely on a single threshold to filter high-confidence predictions. While effective at removing noisy labels, such strict filtering can discard informative ``uncertain'' regions, thus underutilizing unlabeled data. To mitigate this, we propose \textit{fuzzy soft pseudo-labeling} that retains more valuable information from the teacher's predictive distribution.

\subsubsection{Fuzzy Pseudo-Label Generation}

Given a teacher-generated probability map $p^T \in \mathbb{R}^{C \times H \times W}$, where $C$ is the number of classes and $(H, W)$ are spatial dimensions, we select the top-$K$ classes (by predicted probability) at each pixel \citep{qiao2023fuzzy}:
\begin{equation}\label{eq:topk}
    \text{top}K\bigl(p^T_{:,h,w}\bigr) \;=\; \arg \text{sort}_{c \in \{1,\ldots,C\}}\!\bigl(p^T_{c,h,w}\bigr)[\,:\!K].
\end{equation}
The fuzzy pseudo-label distribution $p^{fuzzy}$ is then computed as
\begin{equation}\label{eq:fuzzy}
    p^{\text{fuzzy}}_{c,h,w} =
    \frac{
        p^T_{c,h,w} \cdot \bm{1}_{c \in \operatorname{top}K(p^T_{:,h,w})}
    }{
        \sum\limits_{c' \in \operatorname{top}K(p^T_{:,h,w})} p^T_{c',h,w}
    },
\end{equation}
thus preserving a small set of plausible classes for each location instead of hard-assigning a single class or discarding uncertain pixels entirely. This fuzzy label distribution provides richer supervisory signals, particularly for borderline or potentially misclassified regions, improving the student's ability to learn from uncertain areas.

\subsubsection{Unsupervised Loss with Fuzzy Labels}

We incorporate the fuzzy label distribution into the unsupervised loss $\mathcal{L}_u$ by measuring the KL divergence between the fuzzy pseudo-label $p^{fuzzy}$ and the student's strongly-augmented output $p^S$:
\begin{equation}\label{eq:lu_initial}
    \mathcal{L}_{u} = \frac{1}{N_{valid}} \sum_{h,w} M_{h,w}\,W_{h,w}\,\sum_{c=1}^{C} \,p^{fuzzy}_{c,h,w}\,\log\!\bigl(\frac{p^{fuzzy}_{c,h,w}}{p^S_{c,h,w}}\bigr).
\end{equation}
where
$M_{h,w}$ is a binary mask indicating valid (non-ignored) pixels,
$W_{h,w}$ is a dynamic pixel-level weight tied to the teacher's prediction uncertainty and
$N_{valid}$ is the total number of valid pixels in a training batch.

By combining fuzzy soft labels and consistency regularization, $\mathcal{L}_u$ enforces that the student's predictions on strongly-augmented images align with the teacher's more nuanced soft assignments on weakly-augmented inputs.

\subsection{Uncertainty-Aware Dynamic Weighting}

\subsubsection{Per-Pixel Uncertainty Estimation}

To address these noisy pseudo-labels, we modulate the contribution of each pixel to the unsupervised loss via a pixel-level weight $W_{h,w}$ that inversely depends on normalized entropy. Fig.~\ref{entropy map.png} shows pixel-wise normalized entropy maps for several predictions from the teacher model. Specifically, the normalized entropy $H$ for the teacher's prediction is:
\begin{equation}\label{eq:entropy}
    H\bigl(p^T_{:,h,w}\bigr) \;=\; -\,\frac{1}{\log(C)}\,\sum_{c=1}^{C} p^T_{c,h,w} \,\log\,\bigl(p^T_{c,h,w}\bigr).
\end{equation}
Since $\log(C)$ is the maximum possible entropy (for a uniform distribution), $H\in [0,1]$ conveniently normalizes entropy across different numbers of classes.

\subsubsection{Dynamic Weight Function}

We then define the pixel-wise weight as:
\begin{equation}\label{eq:weight}
    W_{h,w} \;=\; 1 - H\bigl(p^T_{:,h,w}\bigr),
\end{equation}
so that pixels with high uncertainty (close to 1) receive smaller weights, and pixels with low uncertainty (close to 0) have higher weights. This design preserves the benefits of fuzzy labeling while still suppressing contributions from extremely ambiguous teacher predictions.

\subsection{Adaptive Class Rebalancing}

\subsubsection{Mitigating Class-Imbalance Bias}

Semantic segmentation datasets often exhibit significant imbalance (e.g., ``background'' or common object classes can dominate less frequent classes). Under such skew, the model and pseudo-labels inevitably become biased toward majority classes, which can further degrade the quality of minority-class pseudo-labels.

To combat this, we adopt an \textit{adaptive class rebalancing} scheme, computing per-class weights $w_c$ based on the frequency of pseudo-labeled pixels in each training batch. Concretely, for each class $c$, let $F_c$ denote the number of pseudo-labeled pixels. We then compute:
\begin{equation}\label{eq:wc}
    w_c \;=\; \frac{\mathrm{median}(F)}{\,F_c + \epsilon\,},
\end{equation}
where $\mathrm{median}(F)$ is the median of $\{F_c\}_{c=1}^C$ and $\epsilon$ is a small constant (e.g., $10^{-6}$) to avoid division by zero.

\revision{The median-based normaliser in Eq.~\eqref{eq:wc} is drawn from robust statistics, to be more specific: \emph{(i) Robustness of the location estimator.} Because $w_c$ is recomputed at every iteration from the per-batch frequency vector $\{F_c\}$, the normaliser determines how much $w_c$ fluctuates from one batch to the next. The arithmetic mean is inflated by whichever dominant classes (e.g.\ ``background,'' ``road'') happen to appear in a batch; the harmonic mean is dragged toward whichever rare-class frequency is present or absent. The median, by contrast, is insensitive to both tails of the long-tailed segmentation distribution, so $\mathrm{median}(F)$ varies smoothly across batches and yields more consistent per-class gradient scaling over training. \emph{(ii) Bounded magnitude versus inverse frequency.} Pure inverse frequency $w_c = 1/F_c$ is unbounded as $F_c \to 0$, and the $\epsilon$ floor in Eq.~\eqref{eq:wc} only avoids division by zero rather than the steep $1/F_c$ tail; a batch in which a rare class appears with very few pixels therefore produces a single dominant gradient term that destabilises optimisation. Normalising by any robust aggregate of $F$ rescales weights into a bounded range, and among $\{\mathrm{mean}, \mathrm{median}, \mathrm{HM}\}$ the median provides the most stable such anchor. Median-frequency balancing in exactly this form was introduced for semantic segmentation by Eigen and Fergus~\citep{eigen2015predicting} and adopted in SegNet~\citep{badrinarayanan2017segnet}, where it was observed to approximate log-frequency scaling while avoiding the extremes of pure inverse frequency. We empirically validate this choice in Table~\ref{tab:weighting_ablation}, where median-based scaling outperforms inverse, mean, and harmonic-mean alternatives on both Pascal VOC and Cityscapes 1/8 splits.}

\subsection{Loss Reweighting}

We integrate these class-level weights into the unsupervised loss as follows. Letting $c^{\ast}(h,w) = \arg\max_{c} p^{\text{fuzzy}}_{c,h,w}$ denote the dominant fuzzy class at pixel $(h,w)$, we apply $w_{c^{\ast}(h,w)}$ as a per-pixel scalar that multiplies the entire pixel-level KL divergence:
\begin{equation}\label{eq:lu}
\begin{split}
    \mathcal{L}_u
    \;=\; \frac{1}{N_{\text{valid}}} \sum_{h,w} M_{h,w}\,W_{h,w}\,w_{c^{\ast}(h,w)} \\
    \times \sum_{c=1}^C p_{c,h,w}^{\text{fuzzy}} \,\log \!\biggl(\frac{p_{c,h,w}^{\text{fuzzy}}}{p_{c,h,w}^S}\biggr).
\end{split}
\end{equation}
Equivalently, in the pseudocode of Algorithm~\ref{alg.1}, this is implemented as \texttt{w\_cls[lbl]} with \texttt{lbl}~$=$~\texttt{fuzzy.argmax(1)}, broadcast over the spatial KL map. Applying $w_c$ at the pixel level (indexed by the fuzzy argmax class) rather than inside the inner sum over classes preserves the per-pixel KL structure while still rebalancing minority pixels. By boosting contributions of minority-class pixels, the student receives stronger supervision signals for classes that otherwise might be overshadowed. This strategy is conceptually related to reweighting approaches seen in long-tailed classification but tailored here to the specifics of pixel-wise pseudo-label frequencies.

\subsection{Efficient Contrastive Feature Regularization}

\subsubsection{Prototype-Based Contrastive Framework}

Beyond consistency-based training, modern semi-supervised methods often leverage contrastive objectives to better separate feature clusters \citep{chen2020simple,zhou2021c3}. However, straightforward pixel-to-pixel contrast can be computationally expensive due to the sheer number of pixel embeddings per batch. We thus adopt a lightweight prototype-based approach that captures global class-specific representation while minimizing overhead.

For each class $c$, we collect all pixels $i\in P_c$ whose fuzzy pseudo-labels are sufficiently confident (e.g., meeting a confidence threshold or having weight $W_{h,w}$ above some small value). We compute the class prototype $f_c$ as the average (mean) embedding of those pixels:
\begin{equation}\label{eq:prototype}
    f_c \;=\; \frac{1}{|P_c|} \,\sum_{i \in P_c} f_i,
\end{equation}
where $f_i$ is the embedding (e.g., the output of a mid-level feature layer) for pixel $i$.

\subsubsection{Contrastive Loss}

To encourage \textit{intra-class compactness} and better semantic separation, we penalize the cosine distance between each pixel embedding $f_i$ and its corresponding prototype $f_c$:
\begin{equation}\label{eq:contrastive}
    \mathcal{L}_{\text{contrastive}}
    \;=\; \frac{1}{C}\,\sum_{c=1}^{C}\frac{1}{|P_c|}\,\sum_{i \in P_c}\Bigl(1 - \cos(f_i,\,f_c)\Bigr).
\end{equation}
This loss pulls pixel embeddings closer to their class centroid, effectively minimizing within-class variation. Because the prototypes are recomputed at each training step, the model's representations remain updated according to the most recent pseudo-label assignments.

\subsection{Final Training Objective}

\subsubsection{Combined Loss}

Bringing all the pieces together, the overall loss is:
\begin{equation}\label{eq:total}
    \mathcal{L}_{\text{total}}
    = \underbrace{\vphantom{\sum}\mathcal{L}_s}_\text{supervised}
    + \underbrace{\vphantom{\sum}\lambda_u \,\mathcal{L}_u}_\text{unsupervised}
    + \underbrace{\vphantom{\sum}\lambda_c \,\mathcal{L}_{\text{contrastive}}}_\text{contrastive},
\end{equation}
where: \hfill\\
$\mathcal{L}_s$ is the standard cross-entropy loss on labeled data $\mathcal{X}_l$:
          \begin{equation}\label{eq:ls}
              \mathcal{L}_s
              = -\,\frac{1}{|\mathcal{X}_l|} \sum_{(x^l,\,y^l)\in \mathcal{X}_l}
                \sum_{h,w}\,\sum_{c=1}^C y_{c,h,w}^l \,\log \bigl(p_{c,h,w}^S\bigr).
          \end{equation}
$\mathcal{L}_u$ is the KL-divergence-based unsupervised loss using fuzzy soft pseudo-labels and per-pixel weighting. \hfill\\
$\mathcal{L}_{\text{contrastive}}$ is the prototype-based contrastive loss. \hfill\\
$\lambda_u$ and $\lambda_c$ are weighting factors, set to 0.5 and 0.1, respectively in our experiments. $\lambda_u$ and $\lambda_c$ balance the influence of the unsupervised and contrastive losses, respectively.

\subsubsection{Practical Considerations}

\textit{Batch Composition}: We typically sample both labeled and unlabeled examples within each mini-batch to ensure that every update uses both $\mathcal{L}_s$ and $\mathcal{L}_u$.

\textit{Memory and Efficiency}: Prototype-based contrastive regularization is kept efficient by reducing the dimensionality of the features (e.g., via a 128-D projection head) and limiting prototypes to only high-confidence pixels.

\textit{Hyperparameter Tuning}: In practice, $K$ for top-$K$ fuzzy pseudo-labels, the threshold for pixel ``confidence'' in the contrastive branch, and the weighting factors $\lambda_u,\lambda_c$ can be further tuned per dataset for optimal performance.

\section{Experiments}

\subsection{Implementation Details}

Our model is implemented using PyTorch and trained on Pascal VOC and Cityscapes datasets following common semi-supervised segmentation setups~\citep{yang2023revisiting}. We build upon a DeepLabV3+ baseline architecture with ResNet-50 and ResNet-101 backbones~\citep{deng2009imagenet,he2016deep} with dilation rates $\{6,12,18\}$, and adopt standard mean-teacher EMA updates~\citep{sohn2020fixmatch}. Training is performed for 80 epochs for Pascal and 240 epochs for Cityscapes. The initial learning rate is set to $0.001$, following common practice. 

We optimize the model with stochastic gradient descent (SGD) using a momentum factor of $0.9$ and polynomial decay for the learning rate:
\[
    \eta(i) 
    = \eta_0\Bigl(1 - \frac{i}{i_{\max}}\Bigr)^{0.9},
\]
where $\eta_0$ is the initial learning rate, $i$ is the current iteration, and $i_{\max}$ is the total training iterations. Weight decay is set to $10^{-4}$ for regularization. The mean-teacher momentum parameter $\alpha$ is fixed at $0.99$, updating the teacher model as an exponential moving average of the student model's weights.

\subsection{Datasets and Semi-Supervised Setups}

\subsubsection{Pascal VOC (Classic \& Blended)}
We utilize Pascal VOC 2012 dataset \citep{everingham2015pascal} comprising 1,464 high-quality images with extensive labels for training and 1,449
images for validation. We also perform experiments on the Blended PASCAL VOC2012 dataset which is augmented with the Segmentation Boundary Dataset
(SBD) \citep{hariharan2011semantic}, enlarging the training pool to 10,582 images. We also apply the same settings as U2PL \citep{U2PL} for the Blended dataset evaluation. For semi-supervised learning experiments, we adopt the established practice of using fractions \( \{1/16,\,1/8,\,1/4,\,1/2\} \) of the high-quality annotated images as labeled data, treating the remaining images as unlabeled. Evaluation is performed on the official validation set consisting of 1,449 images, and we report the performance in terms of mean Intersection-over-Union (mIoU).

\subsubsection{Cityscapes}
Cityscapes dataset \citep{cordts2016cityscapes} comprises 2,975 training images with fine-grained annotations and 500 validation images, featuring urban scenes labeled across 19 semantic classes. Despite its relatively smaller class space, Cityscapes offers high-resolution annotations (originally $1024\times2048$ pixels), which include detailed labels for small and thin objects such as traffic lights. Model performance is evaluated by calculating the mIoU metric on the provided 500-image validation set.

\subsubsection{Data Augmentation}

Following common practice in semi-supervised segmentation \citep{hu2021semi,ouali2020semi}, we apply weak augmentation (e.g., random scaling, random horizontal flipping) for teacher inference and strong augmentation (e.g., random crops, color jitter, stronger geometric transformations) for student training. Weakly-augmented images yield more stable teacher predictions, while strongly-augmented images regularize the student to learn robust features under varied conditions \citep{ouali2020semi}.

\subsubsection{Hyperparameters} 

\paragraph{Pseudo-Labeling}
For fuzzy pseudo-labeling, we set $K=2$ for Pascal VOC and Cityscapes to balance complexity against memory constraints. A normalized entropy threshold of $0.7$ is used to define high-uncertainty pixels, which are down-weighted via the uncertainty-based weight $W_{h,w}$.

\paragraph{Class Rebalancing} For each mini-batch, we estimate the class-frequency vector $\{F_c\}_{c=1}^C$ over all pixels assigned to classes by the fuzzy pseudo-labels. The final weight $w_c$ is computed with $\epsilon=10^{-6}$.

\paragraph{Contrastive Regularization} For the prototype-based contrastive loss, we project intermediate feature maps into $128$-D embeddings via a lightweight projection head. The hyperparameter $\lambda_c$ in the final loss is set to $0.1$, chosen based on preliminary experiments.

\subsection{Comparisons with State-of-the-Art Methods}

\subsubsection{Quantitative Analysis on Pascal VOC (Classic)}

Table~\ref{tab:pascal} shows the performance of our approach (FARCLUSS) compared to other SOTA methods on Pascal VOC across various labeled data ratios. We provide results for both ResNet-50 and ResNet-101 backbones, indicating the number of labeled examples in parentheses for each split. 
As shown, FARCLUSS consistently achieves competitive or superior mIoU across most labeled ratios, often ranking in the top three methods. Notably, FARCLUSS exceeds UniMatch on every reported cell of the Pascal Classic table by 0.4--1.2 mIoU and reaches CorrMatch's operating point within 0.4 mIoU on every cell, doing so with a single-network design that avoids correspondence-matching overhead.

\begin{table*}[t]
\centering
\caption{Quantitative comparisons of state-of-the-art methods on the Pascal VOC classic setting across different ratios (absolute number of labeled images is written in brackets). The highest values per split are in \textcolor{red}{red}, second highest in \textcolor{green}{green}, and third highest in \textcolor{blue}{blue}.} 
\label{tab:pascal}
\fitpage{
\tabfont
\begin{tabular*}{\textwidth}{@{\extracolsep{\fill}}l c c c c c c c c c c c@{}}
\toprule
\textbf{Pascal SOTAs} & \textbf{Venue} & \multicolumn{5}{c}{\textbf{ResNet-50}} & \multicolumn{4}{c}{\textbf{ResNet-101}} & \textbf{Params} \\
\cmidrule(lr){3-7} \cmidrule(lr){8-11}
 &  & 1/16 & 1/8 & 1/4 & 1/2 & Full & 1/16 & 1/8 & 1/4 & 1/2 &  \\
 &  & (92) & (183) & (366) & (732) & (1464) & (92) & (183) & (366) & (732) &  \\
\midrule
Supervised Only & - & 44.00 & 52.30 & 61.70 & 66.70 & - & 45.1 & 55.3 & 64.8 & 69.7 & 59.5M \\
\midrule
ST++ \citep{ST++}         & CVPR'22   & \textcolor{blue}{72.60} & 74.40 & 75.40 & - & \textcolor{green}{79.10} & 65.2 & 71.0 & 74.6 & 77.3 & 59.5M \\
U$^2$PL \citep{U2PL}       & CVPR'22   & 63.30 & 65.50 & 71.60 & 73.80 & 75.10 & 68.0 & 69.2 & 73.7 & 76.2 & 59.5M \\
PS-MT \citep{liu2022perturbed}          & CVPR'22   & \textcolor{green}{72.80} & \textcolor{blue}{75.70} & 76.02 & 76.64 & \textcolor{red}{80.00} & 65.8 & 69.6 & 76.6 & 78.4 & 59.5M \\
GTA-Seg \citep{GTA-Seg}        & NeurIPS'22 & - & - & - & - & - & 70.0 & 73.2 & 75.6 & 78.4 & 59.5M \\
PCR \citep{PCR}           & NeurIPS'22 & - & - & - & - & - & 70.1 & 74.7 & 77.2 & 78.5 & 59.5M \\
AugSeg \citep{AugSeg}         & CVPR'23   & 64.22 & 72.17 & \textcolor{blue}{76.17} & \textcolor{green}{77.40} & 78.82 & 71.1 & 75.5 & 78.8 & 80.3 & 59.5M \\
Diverse CoT \citep{Diverse_CoT}    & ICCV'23   & - & - & - & - & - & \textcolor{green}{75.7} & 77.7 & \textcolor{red}{80.1} & \textcolor{green}{80.9} & 59.5M \\
ESL \citep{ESL}            & ICCV'23   & 61.74 & 69.50 & 72.63 & 74.69 & 77.11 & 71.0 & 74.0 & 78.1 & 79.5 & 59.5M \\
LogicDiag \citep{LogicDiag}     & ICCV'23   & - & - & - & - & - & 73.3 & 76.7 & 77.9 & 79.4 & 59.5M \\
DAW \citep{DAW}            & NeurIPS'23 & - & - & - & - & - & 74.8 & 77.4 & 79.5 & \textcolor{blue}{80.6} & 59.5M \\
UniMatch \citep{yang2023revisiting} & CVPR'23   & 71.90 & 72.48 & 75.96 & 77.09 & 78.70 & \textcolor{blue}{75.2} & 77.2 & 78.6 & 79.9 & 59.5M \\
DDFP \citep{DDFP}   & CVPR'24   & - & - & - & - & - & 75.0 & \textcolor{blue}{78.0} & \textcolor{green}{79.5} & \textcolor{red}{81.2} & 59.5M \\
CorrMatch \citep{CorrMatch}      & CVPR'24   & - & - & - & - & - & \textcolor{red}{76.4} & \textcolor{red}{78.5} & \textcolor{blue}{79.4} & 80.6 & 59.5M \\
DUEB \citep{smita2025uncertainty}  & WACV'25 & 72.41 & 74.89 & 75.85 & 75.94 & - & 74.1 & 76.7 & 78.0 & 78.1 & 59.5M \\
CW-BASS \citep{tarubinga2025confidence} & IJCNN'25  & \textcolor{green}{72.80} & \textcolor{green}{75.81} & \textcolor{green}{76.20} & \textcolor{blue}{77.15} & - & - & - & - & - & 59.5M \\
\midrule
FARCLUSS       & Ours      & \textcolor{red}{72.90} & \textcolor{red}{76.18} & \textcolor{red}{76.50} & \textcolor{red}{77.69} & \textcolor{blue}{78.90} & \textcolor{red}{76.4} & \textcolor{green}{78.2} & 79.0 & \textcolor{blue}{80.3} & 59.5M \\
\bottomrule
\end{tabular*}
}
\end{table*}

\subsubsection{Quantitative Analysis on Pascal VOC (Blended)}

Table~\ref{tab:pascal_blended_colored} presents a comparison of FARCLUSS against state-of-the-art methods on the Pascal Blended Set using the mIoU (\%) metric. All methods are trained with the `aug` setting, selecting labeled images from the VOC aug training set of 10,582 samples. Results are reported under two training resolutions: 321 and 513. The table distinguishes performance across different labeling ratios (1/16, 1/8, 1/4).
\revision{At the 513 training resolution with ResNet-101, FARCLUSS achieves the top result at 1/16 (78.5 mIoU) and 1/4 (79.8 mIoU), exceeds UniMatch by 0.4--0.6 mIoU across all three splits, and exceeds CorrMatch at 1/16 ($+0.1$) and 1/4 ($+0.2$). On the 1/8 cell, FARCLUSS is within 0.3 mIoU of CorrMatch while using only a single-network pipeline that avoids the correspondence-matching overhead.}

\begin{table*}[t]
\centering
\caption{Quantitative comparisons of state-of-the-art methods on the Pascal Blended set using different training resolutions (321 and 513). All methods are trained using the augmented training split (10,582 images). We evaluate both ResNet-50 and ResNet-101 backbones across three label ratios. The highest values per split are in \textcolor{red}{red}, second highest in \textcolor{green}{green}, and third highest in \textcolor{blue}{blue}.}
\label{tab:pascal_blended_colored}
\fitpage{
\tabfont
\begin{tabular*}{\textwidth}{@{\extracolsep{\fill}}l c c c c c c c c@{}}
\toprule
\textbf{Pascal SOTAs} & \textbf{Train Size} & \textbf{Venue} & \multicolumn{3}{c}{\textbf{ResNet-50}} & \multicolumn{3}{c}{\textbf{ResNet-101}} \\
\cmidrule(lr){4-6} \cmidrule(lr){7-9}
 & & & 1/16 (662) & 1/8 (1323) & 1/4 (2646) & 1/16 (662) & 1/8 (1323) & 1/4 (2646) \\
\midrule
SupOnly Baseline & 321 & - & 61.2 & 67.3 & 70.8 & 65.6 & 70.4 & 72.8 \\
CAC \citep{lai2021semi} & 321 & CVPR'21 & 70.1 & 72.4 & 74.0 & 72.4 & 74.6 & 76.3 \\
ST++ \citep{ST++} & 321 & CVPR'22 & \textcolor{blue}{72.6} & \textcolor{blue}{74.4} & \textcolor{blue}{75.4} & 74.5 & \textcolor{blue}{76.3} & 76.6 \\
UniMatch \citep{yang2023revisiting} & 321 & CVPR'23 & \textcolor{green}{74.5} & \textcolor{blue}{75.8} & \textcolor{green}{76.1} & \textcolor{blue}{76.5} & \textcolor{green}{77.0} & \textcolor{blue}{77.2} \\
CorrMatch \citep{CorrMatch} & 321 & CVPR'24 & - & - & - & \textcolor{red}{77.6} & \textcolor{red}{77.8} & \textcolor{green}{78.3} \\
FARCLUSS & 321 & Ours & \textcolor{red}{74.9} & \textcolor{red}{76.3} & \textcolor{red}{76.4} & \textcolor{green}{77.2} & \textcolor{red}{77.8} & \textcolor{red}{78.4} \\
\midrule
SupOnly Baseline & 513 & - & 62.4 & 68.2 & 72.3 & 67.5 & 71.1 & 74.2 \\
CutMix-Seg \citep{french2019semi} & 513 & BMVC'20 & - & - & - & 71.7 & 75.5 & 77.3 \\
CCT \citep{ouali2020semi} & 513 & CVPR'20 & - & - & - & 71.9 & 73.7 & 76.5 \\
GCT \citep{xia2020guided} & 513 & ECCV'20 & - & - & - & 70.9 & 73.3 & 76.7 \\
CPS \citep{chen2021semisupervised} & 513 & CVPR'21 & 72.0 & 73.7 & 74.9 & 74.5 & 76.4 & 77.7 \\
AEL \citep{hu2021semi} & 513 & NeurIPS'21 & - & - & - & 77.2 & 77.6 & 78.1 \\
FST \citep{du2022learning} & 513 & CVPR'22 & - & - & - & 73.9 & 76.1 & 78.1 \\
ELN \citep{kwon2022semi} & 513 & CVPR'22 & - & - & - & - & 75.1 & 76.6 \\
U$^2$PL \citep{U2PL} & 513 & CVPR'22 & 72.0 & 75.1 & 76.2 & 74.4 & 77.6 & 78.7 \\
PS-MT \citep{liu2022perturbed} & 513 & CVPR'22 & 72.8 & 75.7 & 76.4 & 75.5 & 78.2 & 78.7 \\
AugSeg \citep{AugSeg} & 513 & CVPR'23 & \textcolor{blue}{74.6} & \textcolor{blue}{76.0} & \textcolor{green}{77.2} & 77.0 & 77.3 & 78.9 \\
UniMatch \citep{yang2023revisiting} & 513 & CVPR'23 & \textcolor{green}{75.8} & \textcolor{green}{76.9} & \textcolor{blue}{76.8} & \textcolor{blue}{78.1} & \textcolor{blue}{78.4} & \textcolor{blue}{79.2} \\
CorrMatch \citep{CorrMatch} & 513 & CVPR'24 & - & - & - & \textcolor{green}{78.4} & \textcolor{red}{79.3} & \textcolor{green}{79.6} \\
\midrule
FARCLUSS & 513 & Ours & \textcolor{red}{76.1} & \textcolor{red}{77.3} & \textcolor{red}{77.5} & \textcolor{red}{78.5} & \textcolor{green}{79.0} & \textcolor{red}{79.8} \\
\bottomrule
\end{tabular*}
}
\end{table*}

\subsubsection{Quantitative Analysis on Cityscapes}

We further evaluate on Cityscapes, a more challenging urban-driving dataset with high-resolution images and multiple small classes. Table~\ref{tab:cityscapes_colored} provides comparisons to recent SOTA methods under partial-label settings. 

Despite the greater complexity of Cityscapes (e.g., smaller objects, highly detailed street scenes), our method consistently ranks among the top methods at multiple supervision levels (1/8, 1/4, 1/2). Our fuzzy pseudo-labeling strategy provides fine-grained supervisory signals, beneficial for capturing small classes such as ``rider'' or ``pole.'' Similarly, adaptive class rebalancing helps counteract the long-tail distribution where cars, roads, and buildings dominate. \revision{On ResNet-101, FARCLUSS attains the top 1/2 result at 81.0 mIoU, matching LogicDiag without requiring a logical-diagnosis module. Across the four splits, FARCLUSS exceeds UniMatch by $+0.6$, $+0.6$, $+0.8$, and $+1.5$ mIoU from 1/16 to 1/2, with the largest gap appearing at the most label-rich split, and exceeds CorrMatch at 1/4 ($+0.6$) and 1/2 ($+0.6$). The largest gains over UniMatch concentrate where competing methods invest substantial architectural overhead, indicating that the simpler FARCLUSS pipeline scales to label-rich regimes rather than saturating early.}

\begin{table*}[t]
\centering
\caption{Quantitative comparisons of state-of-the-art methods on the Cityscapes dataset across different ratios (absolute number of labeled images is written in brackets). The highest values per split are in \textcolor{red}{red}, second highest in \textcolor{green}{green}, and third highest in \textcolor{blue}{blue}.}
\label{tab:cityscapes_colored}
\fitpage{
\tabfont
\begin{tabular*}{\textwidth}{@{\extracolsep{\fill}}l c c c c c c c c c c@{}}
\toprule
\textbf{Cityscapes SOTAs} & \textbf{Venue} & \multicolumn{4}{c}{\textbf{ResNet-50}} & \multicolumn{4}{c}{\textbf{ResNet-101}} & \textbf{Params} \\
\cmidrule(lr){3-6} \cmidrule(lr){7-10}
 &  & 1/16 & 1/8 & 1/4 & 1/2 & 1/16 & 1/8 & 1/4 & 1/2 &  \\
 &  & (186) & (372) & (744) & (1488) & (186) & (372) & (744) & (1488) &  \\
\midrule
Supervised Only & - & 63.34 & 68.73 & 74.14 & 76.62 & 66.3 & 72.8 & 75.0 & 78.0 & 59.5M \\
\midrule
U$^2$PL \citep{U2PL}  & CVPR'22   & 69.00 & 73.00 & 76.30 & 78.60 & 74.9 & 76.5 & 78.5 & 79.1 & 59.5M \\
PS-MT \citep{liu2022perturbed}   & CVPR'22   & -     & 75.76 & 76.92 & 77.64 & -    & 76.9 & 77.6 & 79.1 & 59.5M \\
GTA-Seg \citep{GTA-Seg}  & NeurIPS'22 & -     & -     & -     & -     & 69.4 & 72.0 & 76.1 & -    & 59.5M \\
PCR \citep{PCR}            & NeurIPS'22 & -     & -     & -     & -     & 73.4 & 76.3 & 78.4 & 79.1 & 59.5M \\
AugSeg \citep{AugSeg}   & CVPR'23   & 73.70 & 76.40 & \textcolor{red}{78.70} & \textcolor{green}{79.30} & 75.2 & 77.8 & 79.6 & 80.4 & 59.5M \\
Diverse CoT \citep{Diverse_CoT} & ICCV'23   & -     & -     & -     & -     & 75.7 & 77.7 & 78.5 & -    & 59.5M \\
ESL \citep{ESL}      & ICCV'23   & 71.07 & 76.25 & 77.58 & \textcolor{blue}{78.92} & 75.1 & 77.2 & 78.9 & 80.5 & 59.5M \\
LogicDiag \citep{LogicDiag}   & ICCV'23   & -     & -     & -     & -     & 76.8 & \textcolor{red}{78.9} & \textcolor{red}{80.2} & \textcolor{red}{81.0} & 59.5M \\
DAW \citep{DAW}    & NeurIPS'23 & -     & -     & -     & -     & -    & -    & -    & \textcolor{blue}{80.6} & 59.5M \\
UniMatch \citep{yang2023revisiting} & CVPR'23   & \textcolor{green}{75.03} & \textcolor{blue}{76.77} & 77.49 & 78.60 & 76.6 & 77.9 & 79.2 & 79.5 & 59.5M \\
DDFP \citep{DDFP}    & CVPR'24   & -     & -     & -     & -     & \textcolor{blue}{77.1} & \textcolor{blue}{78.2} & \textcolor{blue}{79.9} & \textcolor{green}{80.8} & 59.5M \\
CorrMatch \citep{CorrMatch}  & CVPR'24   & -     & -     & -     & -     & \textcolor{red}{77.3} & \textcolor{green}{78.5} & 79.4 & 80.4 & 59.5M \\
DUEB \citep{smita2025uncertainty}  & WACV'25 & 72.38 & 76.16 & 77.85 & 77.58 & 74.2 & 77.4 & 78.8 & 79.5 & 59.5M \\
CW-BASS \citep{tarubinga2025confidence}  & IJCNN'25         & \textcolor{blue}{75.00} & \textcolor{green}{77.20} & \textcolor{green}{78.43} & -     & -    & -    & -    & -    & 59.5M \\
\midrule
FARCLUSS        & Ours      & \textcolor{red}{75.20} & \textcolor{red}{77.50} & \textcolor{blue}{78.00} & \textcolor{red}{79.60} & \textcolor{green}{77.2} & \textcolor{green}{78.5} & \textcolor{green}{80.0} & \textcolor{red}{81.0} & 59.5M \\
\bottomrule
\end{tabular*}
}
\end{table*}

\subsubsection{Quantitative Analysis on Computational Cost}

\begin{table*}[t]
    \centering
    \caption{Comparison on the Unlabeled data requirements to train all networks of some state-of-the-art semi-supervised semantic segmentation methods on Pascal VOC 2012 and Cityscapes across multiple labeled data partitions. The table reports the number of unlabeled samples utilized during training per epoch.}
    \label{tab:epochperunlabeled_data}
    \fitpage{
    \tabfont
    \begin{tabular*}{\textwidth}{@{\extracolsep{\fill}}lccccccccc@{}}
        \toprule
        \multirow{2}{*}{Method} & \multirow{2}{*}{Networks} 
            & \multicolumn{4}{c}{Pascal VOC 2012 (Classic)} 
            & \multicolumn{4}{c}{Cityscapes} \\
        \cmidrule(lr){3-10}
          &   & 1/16 & 1/8 & 1/4 & 1/2 & 1/16 & 1/8 & 1/4 & 1/2 \\
        \midrule
        CPS \citep{chen2021semisupervised} & 2 
            & 10.5k & 10.5k & 10.5k & 10.5k 
            & 2.9k & 2.9k & 2.9k & 2.9k \\
        ELN \citep{kwon2022semi} & 2
            & 19.8k & 18.5k & 15.8k & 10.5k 
            & 5.5k & 5.2k & 4.4k & 2.9k \\
        PS-MT \citep{liu2022perturbed} & 3 
            & 19.8k & 18.5k & 15.8k & 10.5k 
            & 5.5k & 5.2k & 4.4k & 2.9k \\
        ST++ \citep{ST++} & 4 
            & 14.8k & 13.8k & 11.9k & 7.9k 
            & 4.1k & 3.9k & 3.3k & 2.2k \\
        UniMatch \citep{yang2023revisiting} & 1
            & 9.9k & 9.2k & 7.9k & 5.2k 
            & 2.7k & 2.6k & 2.2k & 1.4k \\
        CW-BASS \citep{tarubinga2025confidence} & 2 
            & 9.9k & 9.2k & 7.9k & 5.2k 
            & 2.7k & 2.6k & 2.2k & 1.4k \\
        \midrule
        \textbf{FARCLUSS (Ours)} & 1
            & \textbf{9.9k} & \textbf{9.2k} & \textbf{7.9k} & \textbf{5.2k} 
            & \textbf{2.7k} & \textbf{2.6k} & \textbf{2.2k} & \textbf{1.4k} \\
        \bottomrule
    \end{tabular*}%
    }
\end{table*}

\begin{figure*}[t]
\centering
\includegraphics[width=\textwidth]{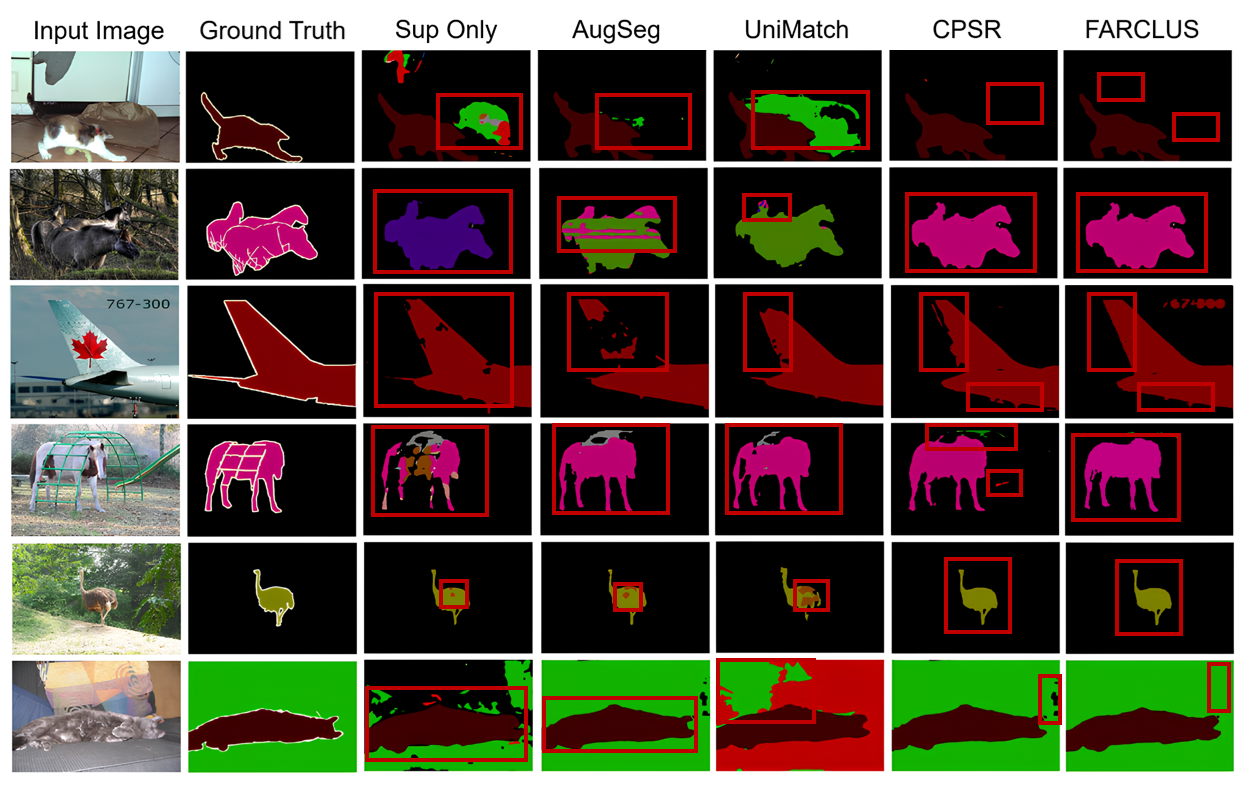}
\caption{Qualitative results of our method, FARCLUSS with other state-of-the-art methods on the PASCAL VOC 2012 dataset. All methods are trained under the Full (1464) setting for validation. \textit{Sup Only} refers to supervised baseline trained only on the corresponding proportion of labeled data. Red rectangles highlight regions where segmentation performance is improved.}
\label{fig3 pascal.png}
\end{figure*}

\begin{figure*}[t]
\centering
\includegraphics[width=\textwidth]{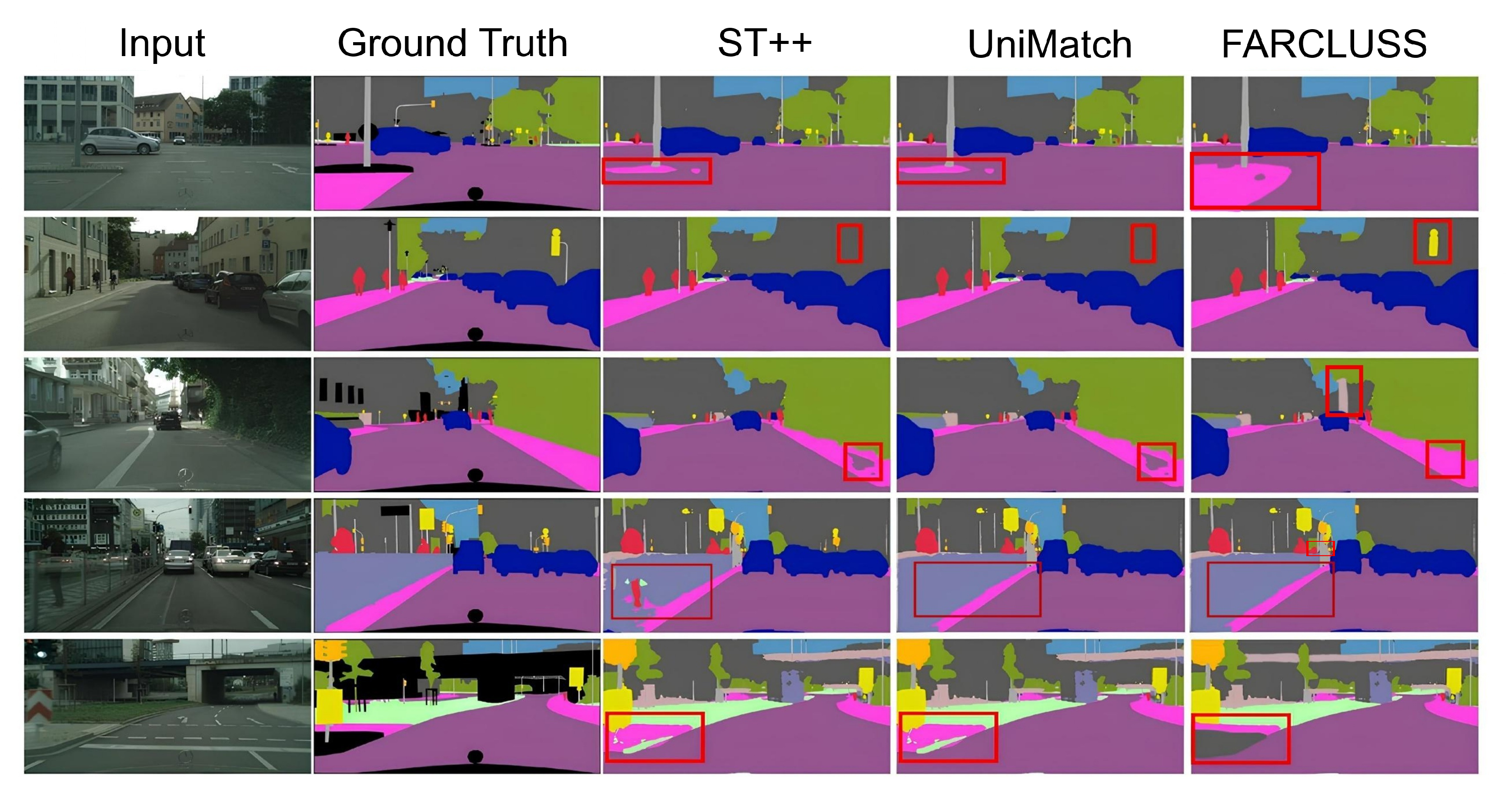}
\caption{Qualitative results of our method, FARCLUSS with other state-of-the-art methods on the Cityscapes dataset. All methods are trained under the 1/8 (12.5\%) setting for validation. Red rectangles highlight regions where segmentation performance is improved.}
\label{cityscapes.png}
\end{figure*}

Efficient utilization of unlabeled data in semi-supervised semantic segmentation critically influences performance, computational resources, and training stability. Comparative analysis with state-of-the-art methods show that FARCLUSS significantly reduces reliance on unlabeled data, matching the minimal requirements observed in recent approaches such as UniMatch \citep{yang2023revisiting} and CW-BASS \citep{tarubinga2025confidence} as shown in Table~\ref{tab:epochperunlabeled_data}.

Earlier methods, including CPS \citep{chen2021semisupervised}, PS-MT \citep{liu2022perturbed}, and ST++ \citep{ST++}, require extensive unlabeled datasets due to inherent architectural and training complexities. For example, CPS \citep{chen2021semisupervised} employs dual networks generating mutual pseudo-labels, thus demanding extensive unlabeled data to sustain effective mutual supervision. For example, it processes 10.5k and 2.9k unlabeled images per epoch on Pascal VOC and Cityscapes, respectively. Its reliance on a fixed quantity of unlabeled data, irrespective of the labeled fraction, highlights the substantial data requirement for stable training outcomes.

Similarly, PS-MT \citep{liu2022perturbed} incorporates multiple perturbation methods (input, feature, and network-level) along with stringent confidence thresholds within three networks. Such complexities necessitate extensive unlabeled datasets to maintain robust consistency learning and reduce the impact of inaccurate pseudo-labels. It requires double or triple the unlabeled data and about 350--550 epochs on Cityscapes for model convergence.
ST++ \citep{ST++} uses self-training combined with selective retraining and robust data augmentations. These strategies substantially increase unlabeled data requirements (14.8k on Pascal, 4.1k on Cityscapes per epoch), especially under scenarios with low labeled data fractions, to ensure training stability.

UniMatch \citep{yang2023revisiting} adopts a streamlined weak-to-strong consistency framework using a shared student-teacher architecture. This reduces redundancy and enhances efficiency by utilizing consistency regularization without extensive pseudo-labeling, thereby effectively decreasing the required volume of unlabeled data. 
CW-BASS \citep{tarubinga2025confidence} introduces confidence-weighted losses and boundary-aware modules targeting confirmation bias and boundary ambiguities within a dual network using fewer augmentations. These enhancements allow the model to emphasize high-confidence regions, significantly lowering the necessity for large unlabeled datasets (up to 9.9k on Pascal, 2.7k on Cityscapes). 

FARCLUSS employs data-efficient techniques similar to UniMatch and CW-BASS, reducing reliance on large unlabeled datasets. The approach uses a shared student-teacher network with fuzzy pseudo-labeling to capture prediction uncertainty, while uncertainty-based weighting attenuates noise from erroneous pseudo-labels and stabilizes training. Adaptive class rebalancing addresses underrepresented classes, reducing the required dataset size for rare class representation and facilitating convergence without extensive unlabeled data-driven exploration. Consistent with efficient prior state-of-the-art methods, the model is trained for 80 epochs on Pascal and 240 epochs on Cityscapes. In this regard, our method enhances data efficiency and reduces computational cost significantly. \revision{Beyond the unlabeled-data efficiency reported in Table~\ref{tab:epochperunlabeled_data}, FARCLUSS achieves its results with a strictly simpler pipeline than the methods it ties on benchmark tables: CorrMatch requires dense correspondence matching between augmented views, LogicDiag requires logical-diagnosis modules, U$^2$PL requires pixel-level pairwise contrastive sampling with unreliable negatives, and ReCo requires region-level pairwise sampling with active negative mining. FARCLUSS requires only fuzzy top-$K$ pseudo-labelling, normalised-entropy weighting, and lightweight prototype-based contrastive regularization (analysed in Section~\ref{sec:contrastive_ablations}; see Table~\ref{tab:contrastive_comparison}), all bolted onto a single-network mean-teacher pipeline.}

\subsubsection{Qualitative Analysis}

Figs.~\ref{fig3 pascal.png} and~\ref{cityscapes.png} show visual comparisons of segmentation outputs on Pascal VOC and Cityscapes. Our FARCLUSS model demonstrates clearer boundaries and improved recognition of minority classes (e.g., objects occupying a small fraction of the image). Fuzzy pseudo-labeling helps the student learn more stable decision boundaries around ambiguous regions, while the prototype-based contrastive learning better separates class embeddings (see Table~\ref{tab:miou_results}).

\subsection{Ablation Study}

To validate the contributions of each component in our framework, we conduct a comprehensive ablation study on the Pascal VOC (1/8 labeled data) and Cityscapes (1/8 labeled data) datasets using a ResNet-101 backbone. We systematically remove individual components while keeping others intact and analyze their impact on segmentation performance (mIoU). We evaluate the four key components of our method: Fuzzy Pseudo-Labeling ($\mathcal{L}_f$), Uncertainty-Aware Weighting ($W_{h,w}$), Adaptive Class Rebalancing ($w_c$), and Contrastive Loss ($\mathcal{L}_c$). The results are summarized in Table~\ref{tab:ablation}.

\begin{table}[t]
\centering
\caption{Ablation study on Pascal VOC 1/8 and Cityscapes 1/8 splits using ResNet-101. Each row shows which components are active (\checkmark). The mIoU reflects the performance of each configuration. Components: 
$\mathcal{L}_f$ = fuzzy pseudo-labeling, $W_{h,w}$ = uncertainty-based pixel-wise weighting, $w_c$ = adaptive class rebalancing, $\mathcal{L}_c$ = prototype contrastive loss.}
\label{tab:ablation}
 \fitcol{%
\tabfont
\begin{tabular*}{\columnwidth}{@{\extracolsep{\fill}}lcccccc@{}}
\toprule
\textbf{Variant} & $\mathcal{L}_f$ & $W_{h,w}$ & $w_c$ & $\mathcal{L}_c$ & \textbf{Pascal 1/8} & \textbf{Cityscapes 1/8} \\
\midrule
Baseline ($\mathcal{L}_s$ only) & \texttimes & \texttimes & \texttimes & \texttimes & 55.3 & 72.8 \\
w/o $\mathcal{L}_f$ & \texttimes & \checkmark & \checkmark & \checkmark & 72.0 & 73.5 \\
w/o $W_{h,w}$ & \checkmark & \texttimes & \checkmark & \checkmark & 75.0 & 74.7 \\
w/o $w_c$ & \checkmark & \checkmark & \texttimes & \checkmark & 76.5 & 75.9 \\
w/o $\mathcal{L}_c$ & \checkmark & \checkmark & \checkmark & \texttimes & 77.5 & 78.1 \\
\midrule
\textbf{Full Method} & \checkmark & \checkmark & \checkmark & \checkmark & \textbf{78.2} & \textbf{78.5} \\
\bottomrule
\end{tabular*}%
}
\end{table}

\paragraph{Synergistic Effects} The combined model exhibits super-additive improvements, reaching 78.2 mIoU on Pascal and 78.5 on Cityscapes (1/8 splits). Notably, fuzzy labeling and entropy weighting jointly allow learning from uncertain regions without degrading performance. Class rebalancing corrects label bias, and contrastive regularization further refines representation quality. The interaction between these modules leads to improved generalization, greater training stability, and more balanced performance across the class spectrum.

\subsubsection{Component Analysis}

\paragraph{Fuzzy Pseudo-Labeling (L$_f$)} This module delivers the most substantial drop (-6.2 mIoU on Pascal), confirming its role in retaining informative supervision from ambiguous regions. By preserving top-$K$ class probabilities, fuzzy labels reduce confirmation bias and mitigate the overconfidence common in hard pseudo-labels. This mechanism aligns with entropy regularization and label smoothing, offering a continuous signal that better captures uncertainty, especially at object boundaries and for underrepresented classes.

\paragraph{Uncertainty-Aware Weighting ($W_{h,w}$)} Weighting pixel contributions using normalized entropy improves robustness by down-weighting noisy pseudo-labels, particularly in the early training stages. This acts as a soft curriculum, allowing the model to focus on high-confidence regions before incorporating noisier predictions. Removing this component results in a $\sim$3.2 mIoU drop on Pascal (and $\sim$3.8 on Cityscapes), indicating its importance in suppressing confirmation errors and ensuring stability.

\paragraph{Adaptive Class Rebalancing ($w_c$)} This component addresses the skewed distribution of pseudo-labels by adjusting loss contributions per pixel based on the pseudo-labelled class frequency. Removing $w_c$ causes a 1.7 mIoU drop on Pascal (78.2 $\rightarrow$ 76.5) and a 2.6 mIoU drop on Cityscapes (78.5 $\rightarrow$ 75.9) at the 1/8 split, with the larger Cityscapes effect consistent with its more pronounced long-tail distribution (e.g.\ ``road'' and ``building'' dominating ``rider,'' ``pole''). The approach is consistent with long-tail-learning reweighting strategies such as DARS \citep{cho2021dars}.

\paragraph{Contrastive Loss ($\mathcal{L}_c$)} Though its isolated effect is more modest (0.4--0.7 mIoU), the contrastive loss enhances feature discriminability through intra-class compactness and inter-class separation. Our lightweight, prototype-based implementation avoids the computational cost of pairwise sampling and proves especially effective in disambiguating semantically similar regions.

\revision{
\subsubsection{Class Weighting Strategy Ablation}
\label{sec:weighting_ablation}
To justify the choice of median-based frequency scaling in Eq.~\eqref{eq:wc}, we compare four class weighting strategies: (i) \textit{inverse frequency} ($w_c = 1/F_c$), (ii) \textit{mean-based} ($w_c = \mathrm{mean}(F)/F_c$), (iii) \textit{harmonic mean-based} ($w_c = \mathrm{HM}(F)/F_c$), and (iv) \textit{median-based} ($w_c = \mathrm{median}(F)/F_c$, ours). Results on Pascal VOC and Cityscapes 1/8 splits with ResNet-101 are reported in Table~\ref{tab:weighting_ablation}.
}

\begin{table}[t]
\centering
\revision{
\caption{Ablation on class weighting strategies (Eq.~\eqref{eq:wc}) using ResNet-101 on 1/8 splits. mIoU (\%) is reported.}
\label{tab:weighting_ablation}
\tabfont
\begin{tabular*}{\columnwidth}{@{\extracolsep{\fill}}lcc@{}}
\toprule
\textbf{Weighting Strategy} & \textbf{Pascal 1/8} & \textbf{Cityscapes 1/8} \\
\midrule
No rebalancing (uniform) & 76.5 & 75.9 \\
Inverse frequency ($1/F_c$) & 76.1 & 75.4 \\
Mean-based & 77.4 & 77.1 \\
Harmonic mean-based & 76.8 & 76.5 \\
\textbf{Median-based (Ours)} & \textbf{78.2} & \textbf{78.5} \\
\bottomrule
\end{tabular*}
}
\end{table}

\revision{As shown, inverse frequency weighting assigns unbounded weights to extremely rare classes, causing gradient instability and the worst result on both benchmarks (76.1 / 75.4 mIoU). Harmonic-mean-based weighting recovers some of this loss (76.8 / 76.5) but the harmonic mean is itself sensitive to the rare frequencies it is meant to up-weight, so the per-batch numerator fluctuates more than the median and the effective rebalancing is muted. Mean-based weighting recovers further (77.4 / 77.1) but $\mathrm{mean}(F)$ is inflated by the dominant tail and varies batch-to-batch with whichever majority classes happen to dominate. Median-based scaling delivers the best result on both benchmarks (78.2 / 78.5) by providing a batch-level location estimate that is insensitive to either tail of the class-frequency distribution. 
}

\revision{
\subsubsection{Sensitivity Analysis: Top-$K$ Fuzzy Labeling vs.\ Fixed Thresholding}
\label{sec:topk_sensitivity}
A central design choice in FARCLUSS is the use of top-$K$ fuzzy pseudo-labeling rather than conventional fixed-threshold filtering. To empirically justify this choice, we compare: (i) top-$K$ selection with $K \in \{1, 2, 3, 5\}$, and (ii) fixed confidence thresholds $\tau \in \{0.95, 0.90, 0.80\}$, where pixels with $\max_c p^T_{c,h,w} < \tau$ are discarded entirely. Results are reported in Table~\ref{tab:topk_vs_threshold}.
}

\begin{table}[t]
\centering
\revision{
\caption{Top-$K$ fuzzy labeling vs.\ fixed-threshold filtering on Pascal VOC and Cityscapes 1/8 splits using ResNet-101. ``Retention'' denotes the average percentage of unlabeled pixels retained for training.}
\label{tab:topk_vs_threshold}
\fitcol{%
\tabfont
\begin{tabular*}{\columnwidth}{@{\extracolsep{\fill}}lccc@{}}
\toprule
\textbf{Strategy} & \textbf{Pascal 1/8} & \textbf{Cityscapes 1/8} & \textbf{Retention (\%)} \\
\midrule
\multicolumn{4}{l}{\textit{Fixed confidence threshold}} \\
\quad $\tau = 0.95$ & 76.8 & 77.0 & 63.2 \\
\quad $\tau = 0.90$ & 77.3 & 77.6 & 76.5 \\
\quad $\tau = 0.80$ & 76.5 & 76.9 & 88.1 \\
\midrule
\multicolumn{4}{l}{\textit{Top-$K$ fuzzy pseudo-labeling (Ours)}} \\
\quad $K = 1$ (hard) & 77.0 & 77.4 & 100.0 \\
\quad $K = 2$ (default) & \textbf{78.2} & \textbf{78.5} & 100.0 \\
\quad $K = 3$ & 77.8 & 78.1 & 100.0 \\
\quad $K = 5$ & 77.2 & 77.5 & 100.0 \\
\bottomrule
\end{tabular*}%
}
}
\end{table}

\revision{The key distinction between the two strategies is that fixed thresholding operates in probability space, discarding all pixels below a confidence cutoff, whereas top-$K$ selection operates in rank space, retaining all pixels but limiting the number of candidate classes per pixel. Consequently, top-$K$ achieves 100\% pixel retention regardless of $K$, whereas fixed thresholds discard increasingly large fractions of the training signal as $\tau$ increases. At $\tau = 0.95$, a substantial portion of boundary and ambiguous pixels are eliminated, forfeiting valuable supervisory signals. At $\tau = 0.80$, more pixels are retained but noisy predictions degrade performance. Top-$K$ with $K=2$ strikes the optimal balance since it preserves the full spatial extent of the unlabeled data while constraining the label distribution to the two most plausible classes, effectively capturing boundary ambiguity without introducing excessive noise from unlikely classes ($K=5$) or collapsing to hard assignment ($K=1$). These results confirm that our top-$K$ scheme represents a fundamentally different and superior strategy compared to fixed-threshold filtering, as it decouples the retention decision from the confidence magnitude. This finding is consistent with recent soft pseudo-label filtering strategies such as SoftMatch~\citep{chen2023softmatch} and FreeMatch~\citep{wang2023freematch}, which similarly advocate preserving low-confidence predictions through adaptive weighting rather than hard elimination.}

\subsection{Loss-Convergence Dynamics}
\label{sec:Loss-Convergence}

\begin{figure}[b]
    \centering
    \includegraphics[width=\linewidth]{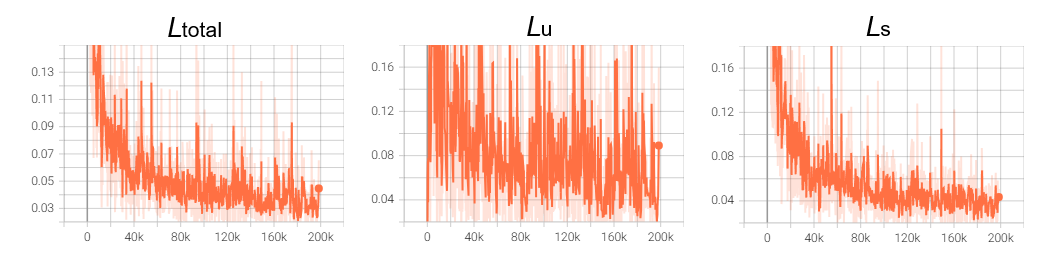}
    \caption{Training trajectories of the total loss
\(\mathcal{L}_{\mathrm{total}}\) alongside its supervised component, \(\mathcal{L}_{s}\) and unsupervised component, \(\mathcal{L}_{u}\) over the full 80-epoch Pascal VOC training schedule on the 1/16 labeled split (Section~\ref{sec:Loss-Convergence}). For consistency with prior loss-curve visualisations, the $x$-axis is reported as iteration count rather than epoch index.}
    \label{fig:Loss-Convergence}
\end{figure}

\paragraph{Total Loss}
Fig.~\ref{fig:Loss-Convergence} shows a monotonic descent from $\approx0.14$ to $\approx0.03$.  
Because $\lambda_{c}\ll\lambda_{u}$, curvature is dominated by $\mathcal{L}_{s}+\lambda_{u}\mathcal{L}_{u}$; spikes map directly to bursts in $\mathcal{L}_{u}$.  
The trend confirms stable optimization and a negligible but beneficial compactness term, consistent with findings in regional contrastive learning \citep{liu2022reco}.

\paragraph{Supervised Loss $\mathcal{L}_{s}$}
A rapid fall within the first $4\!\times\!10^{4}$ updates reflects quick memorization of the small labeled subset.
The plateau near $0.04$ indicates saturation once boundary-level errors dominate; labeled pixels contribute only $1/16$ of batch gradients, limiting further reduction.
Low variance follows from fixed one-hot labels and absence of entropy weighting.

\paragraph{Unsupervised Loss $\mathcal{L}_{u}$}
An initial value $>0.16$ arises from wide-support pseudo-labels produced by an immature EMA teacher.  
High variance persists because pixel weights $W_{h,w}=1-H(p_T)$ and class weights $w_c$ are recomputed each iteration, a behavior similarly reported in U$^{2}$PL \citep{U2PL}.
Downward drift to $\approx0.06$ evidences teacher refinement and entropy reduction; late-phase spikes coincide with adaptive boosting of minority classes, matching the goals of class rebalancing modules.

\paragraph{Consistency with Prior Work}
Fuzzy pseudo-labeling preserves gradient signal in ambiguous regions, avoiding the loss collapse typical of hard-threshold methods such as FixMatch \citep{sohn2020fixmatch} or CPS \citep{chen2021semisupervised}.
Entropy weighting keeps variance bounded; the prototype-based contrastive regularization supplies tighter intra-class feature clusters with marginal numerical load.
Observed dynamics replicate these effects, verifying correct implementation of FARCLUSS under extreme label scarcity.

\revision{\subsection{Entropy Calibration and Reliability Analysis}\label{sec:entropy_calibration}}

\revision{A potential concern with entropy-based weighting (Eq.~5--6) is that an overconfident teacher may produce artificially low-entropy predictions on incorrect regions, thereby assigning high weights to erroneous pseudo-labels. We address this concern by directly measuring the Expected Calibration Error (ECE)~\citep{guo2017calibration} of the EMA teacher on unlabeled data throughout training. The same measurement supports two conclusions: it provides a quantitative check that calibration remains stable (and improves) over the training schedule, and it provides empirical evidence that the fuzzy top-$K$ target distribution structurally attenuates overconfidence by distributing probability mass across plausible classes rather than concentrating it on a single argmax.}

\revision{
ECE measures the average discrepancy between predicted confidence and actual accuracy across confidence bins:}
\begin{equation}\label{eq:ece}
    \mathrm{ECE} = \sum_{m=1}^{M} \frac{|B_m|}{N} \bigl| \mathrm{acc}(B_m) - \mathrm{conf}(B_m) \bigr|,
\end{equation}
\revision{where $B_m$ denotes the set of pixels in confidence bin $m$, $\mathrm{acc}(B_m)$ is the accuracy within that bin, and $\mathrm{conf}(B_m)$ is the mean confidence. We use $M=15$ bins following standard practice.}

\begin{table}[t]
\centering
\revision{
\caption{Expected Calibration Error (ECE, \%) of the EMA teacher on Pascal VOC 1/8 (ResNet-101) at different training stages. Lower is better.}
\label{tab:ece}
\tabfont
\begin{tabular*}{\columnwidth}{@{\extracolsep{\fill}}lcccc@{}}
\toprule
\textbf{Method} & \textbf{50k} & \textbf{100k} & \textbf{150k} & \textbf{200k} \\
\midrule
UniMatch (teacher) & 5.82 & 5.21 & 4.93 & 4.68 \\
FARCLUSS (teacher) & \textbf{4.57} & \textbf{4.12} & \textbf{3.84} & \textbf{3.61} \\
\bottomrule
\end{tabular*}
}
\end{table}

\revision{
Table~\ref{tab:ece} shows that our EMA teacher maintains low ECE throughout training and, importantly, ECE does not increase at later stages, which would be the signature of overfitting-induced overconfidence. FARCLUSS's teacher achieves consistently lower ECE than UniMatch's teacher at every checkpoint, and the within-run trend is monotone decreasing (4.57 $\rightarrow$ 4.12 $\rightarrow$ 3.84 $\rightarrow$ 3.61). We attribute this to the combination of fuzzy pseudo-labelling and entropy-based weighting: distributing probability mass across the top-$K$ classes rather than concentrating it on a single argmax acts as a label-smoothing-style regulariser that discourages peaked posteriors in ambiguous regions.
}

\revision{
\subsection{Contrastive Regularization Ablations}
\label{sec:contrastive_ablations}
We conduct detailed ablations on the prototype-based contrastive regularization (Eqs.~\eqref{eq:prototype}--\eqref{eq:contrastive}) to validate key design choices. 
}
\revision{
\subsubsection{Confidence Threshold for Prototype Selection}
Prototypes are computed from pixels whose uncertainty-based weight $W_{h,w}$ exceeds a threshold $\tau_{\text{proto}}$. We examine $\tau_{\text{proto}} \in \{0.3, 0.5, 0.7, 0.8, 0.9, 0.95\}$ on Pascal VOC 1/8 (ResNet-101) in Table~\ref{tab:proto_threshold}.
}

\begin{table}[t]
\centering
\revision{
\caption{Ablation on prototype confidence threshold $\tau_{\text{proto}}$ on Pascal VOC 1/8 (ResNet-101).}
\label{tab:proto_threshold}
\tabfont
\begin{tabular*}{\columnwidth}{@{\extracolsep{\fill}}ccc@{}}
\toprule
$\tau_{\text{proto}}$ & \textbf{mIoU (\%)} & \textbf{Pixels Used (\%)} \\
\midrule
0.3 & 77.5 & 82.6 \\
0.5 (default) & \textbf{78.2} & 68.4 \\
0.7 & 77.9 & 53.7 \\
0.8 & 77.6 & 44.2 \\
0.9 & 77.1 & 31.5 \\
0.95 & 76.7 & 22.8 \\
\bottomrule
\end{tabular*}
}
\end{table}

\begin{figure}[t]
\centering
\revision{
\begin{tikzpicture}
\begin{axis}[
  width=0.92\columnwidth, height=5.6cm,
  xlabel={\footnotesize Prototype confidence threshold $\tau_{\text{proto}}$},
  ylabel={\footnotesize mIoU (\%)},
  xmin=0.25, xmax=1.0,
  ymin=76.2, ymax=78.8,
  ytick={76.5, 77.0, 77.5, 78.0, 78.5},
  axis y line*=left,
  xtick={0.3, 0.5, 0.7, 0.8, 0.9, 0.95},
  xticklabel style={font=\scriptsize, rotate=45, anchor=east},
  yticklabel style={font=\scriptsize},
  grid=major, grid style={dashed, gray!25},
  legend style={at={(0.03,0.04)}, anchor=south west, font=\scriptsize, draw=none, fill=white, fill opacity=0.9, text opacity=1},
  every axis plot/.append style={very thick},
]
\addplot[
  mark=*, color=blue!75!black, mark size=2.6pt,
  nodes near coords,
  nodes near coords align={vertical},
  every node near coord/.append style={
    font=\scriptsize, color=blue!80!black, yshift=3pt,
    /pgf/number format/fixed, /pgf/number format/fixed zerofill, /pgf/number format/precision=1
  }
] coordinates {
  (0.3, 77.5) (0.5, 78.2) (0.7, 77.9) (0.8, 77.6) (0.9, 77.1) (0.95, 76.7)
};
\addlegendentry{mIoU}
\addplot[mark=none, gray!50, dotted, thick] coordinates {(0.25, 78.2) (1.0, 78.2)};
\end{axis}
\begin{axis}[
  width=0.92\columnwidth, height=5.6cm,
  ylabel={\footnotesize Pixels used (\%)},
  xmin=0.25, xmax=1.0,
  ymin=15, ymax=95,
  ytick={20, 40, 60, 80},
  axis y line*=right, axis x line=none,
  yticklabel style={font=\scriptsize},
  legend style={at={(0.03,0.20)}, anchor=south west, font=\scriptsize, draw=none, fill=white, fill opacity=0.9, text opacity=1},
  every axis plot/.append style={very thick, dashed},
]
\addplot[mark=square*, color=orange!85!black, mark size=2.2pt] coordinates {
  (0.3, 82.6) (0.5, 68.4) (0.7, 53.7) (0.8, 44.2) (0.9, 31.5) (0.95, 22.8)
};
\addlegendentry{Pixels used}
\end{axis}
\end{tikzpicture}
\caption{Prototype confidence threshold $\tau_{\text{proto}}$ vs.\ mIoU (blue, left axis) and pixel retention (orange, right axis) on Pascal VOC 1/8 (ResNet-101). Numeric labels on every data point. The mIoU curve is unimodal with peak 78.2 at $\tau_{\text{proto}}=0.5$ (dotted reference line); the 1.5 mIoU swing from $\tau_{\text{proto}}=0.5$ to $\tau_{\text{proto}}=0.95$ tracks the steep drop in pixels admitted to the prototype.}
\label{fig:proto_threshold}
}
\end{figure}

\revision{
Fig.~\ref{fig:proto_threshold} visualises the unimodal relationship in Table~\ref{tab:proto_threshold}. Low thresholds ($\tau_{\text{proto}} \leq 0.3$) admit noisy pixels into prototypes, diluting class-specific representations. Very high thresholds ($\tau_{\text{proto}} \geq 0.9$) restrict prototypes to a small fraction of pixels, reducing representativeness and yielding unstable centroids for rare classes. The default $\tau_{\text{proto}} = 0.5$ achieves the best trade-off, retaining a sufficient pool of reliable pixels while excluding the most uncertain predictions.
}

\revision{
\subsubsection{Prototype Embedding Dimension}
We vary the projection head output dimension $d \in \{64, 128, 256, 512\}$ and report results in Table~\ref{tab:proto_dim}.
}

\begin{table}[t]
\centering
\revision{
\caption{Ablation on prototype embedding dimension $d$ on Pascal VOC 1/8 (ResNet-101). GPU memory is measured per batch.}
\label{tab:proto_dim}
\tabfont
\begin{tabular*}{\columnwidth}{@{\extracolsep{\fill}}ccc@{}}
\toprule
\textbf{Dimension $d$} & \textbf{mIoU (\%)} & \textbf{GPU Memory (GB)} \\
\midrule
64 & 77.8 & 8.2 \\
128 (default) & \textbf{78.2} & 8.5 \\
256 & 78.1 & 9.1 \\
512 & 78.0 & 10.4 \\
\bottomrule
\end{tabular*}
}
\end{table}

\revision{
Dimension $d=128$ achieves optimal performance with modest memory overhead. Reducing to $d=64$ slightly degrades mIoU due to insufficient representational capacity, while increasing to $d=256$ or $d=512$ provides no benefit and increases memory consumption, confirming that the 128-D projection head is well-suited for the prototype contrastive objective.
}

\revision{
\subsubsection{Comparison with Pairwise Contrastive Baselines}
\label{sec:pairwise_comparison}
We conduct additional comparisons with representative pairwise contrastive learning baselines. Specifically, we compare our method with ReCo~\citep{liu2022reco} and U$^2$PL~\citep{U2PL}. ReCo performs region-level pixel-pair contrastive learning with active negative sampling, while U$^2$PL exploits unreliable pseudo-labels as negative samples in a pixel-level contrastive objective.
For a fair and controlled comparison, each contrastive objective is integrated into the proposed FARCLUSS framework by replacing only $\mathcal{L}_{\text{contrastive}}$. All other components, including the network architecture, data splits, training protocol, and remaining hyperparameters, are kept unchanged. We report the segmentation performance in terms of mIoU, as well as the training time per iteration and peak GPU memory usage, in Table~\ref{tab:contrastive_comparison}. This evaluation enables a direct comparison of both accuracy and computational efficiency across different contrastive regularization strategies.
}

\begin{table}[t]
\centering
\revision{
\caption{Comparison of contrastive regularization strategies integrated into FARCLUSS on Pascal VOC 1/8 (ResNet-101). Time and memory are measured on a single NVIDIA GPU.}
\label{tab:contrastive_comparison}
\fitcol{%
\tabfont\setlength{\tabcolsep}{2pt}
\begin{tabular*}{\columnwidth}{@{\extracolsep{\fill}}lccc@{}}
\toprule
\textbf{Contrastive Method} & \textbf{mIoU (\%)} & \textbf{Time/iter (ms)} & \textbf{Memory (GB)} \\
\midrule
No contrastive loss & 77.5 & 452 & 8.0 \\
ReCo~\citep{liu2022reco} (pairwise) & 78.0 & 618 & 11.3 \\
U$^2$PL~\citep{U2PL} (pairwise) & 77.8 & 584 & 10.6 \\
\textbf{Prototype-based (Ours)} & \textbf{78.2} & \textbf{467} & \textbf{8.5} \\
\bottomrule
\end{tabular*}%
}
}
\end{table}

\revision{
Our prototype-based contrastive loss achieves comparable or superior mIoU to pairwise baselines while requiring significantly less computation and memory. ReCo's region-level pairwise comparisons introduce substantial overhead due to the need to compute and store dense similarity matrices. U$^2$PL's pixel-level contrastive objective similarly scales poorly with spatial resolution. In contrast, our approach computes only $C$ class prototypes per batch and evaluates cosine similarity between each pixel and its corresponding prototype, yielding $\mathcal{O}(N \cdot d)$ complexity compared to $\mathcal{O}(N^2 \cdot d)$ for pairwise methods, where $N$ is the number of selected pixels and $d$ is the embedding dimension. 
}

\subsection{Further Quantitative Analysis on PASCAL VOC 2012 Classes' Results}
\label{sec:perclass_results}

\begin{table*}[t]
\centering
\caption{Quantitative results on PASCAL VOC 2012 per class categories of our method, FARCLUSS, and the Supervised Baseline, CPCL \citep{fan2023conservative} \& DUEB \citep{smita2025uncertainty} methods using ResNet-50. Categories: \textbf{Animal} (Bird, Cat, Cow, Dog, Horse, Sheep), \textbf{Vehicle} (Aeroplane, Bicycle, Boat, Bus, Car, Motorbike, Train), \textbf{Indoor} (Bottle, Chair, Dining Table, Potted Plant, Sofa, Monitor), \textbf{Person}, and \textbf{Background}.}

\tabfont
\begin{tabular*}{\textwidth}{@{\extracolsep{\fill}}lcccccccccccc@{}}
\toprule
\textbf{Methods} & \multicolumn{4}{c}{\textbf{Avg mIoU}} & \multicolumn{4}{c}{\textbf{Animal}} & \multicolumn{4}{c}{\textbf{Vehicle}} \\
\cmidrule(lr){2-5} \cmidrule(lr){6-9} \cmidrule(lr){10-13}
 & 1/2 & 1/4 & 1/8 & 1/16 & 1/2 & 1/4 & 1/8 & 1/16 & 1/2 & 1/4 & 1/8 & 1/16 \\
\midrule
Supervised & 74.05 & 71.66 & 67.16 & 62.00 & 84.45 & 78.98 & 73.47 & 67.65 & 76.90 & 74.31 & 71.94 & 67.88 \\
CPCL \citep{fan2023conservative} & 75.30 & 74.58 & 73.74 & 71.66 & 86.63 & 85.39 & 84.24 & 82.73 & 77.23 & 76.62 & 76.53 & 76.24 \\
DUEB \citep{smita2025uncertainty} & 75.94 & 75.85 & 74.89 & 72.41 & 87.08 & 86.69 & 85.47 & 84.66 & 78.25 & 78.60 & 78.75 & 77.09 \\
\textbf{FARCLUSS (Ours)} & \textbf{77.69} & \textbf{76.50} & \textbf{76.18} & \textbf{72.90} & \textbf{88.25} & \textbf{87.89} & \textbf{86.50} & \textbf{85.73} & \textbf{79.86} & \textbf{80.17} & \textbf{80.02} & \textbf{78.93} \\
\bottomrule
\end{tabular*}

\vspace{0.4em} 

\tabfont
\begin{tabular*}{\textwidth}{@{\extracolsep{\fill}}lcccccccccccc@{}}
\toprule
\textbf{Methods} & \multicolumn{4}{c}{\textbf{Indoor}} & \multicolumn{4}{c}{\textbf{Person}} & \multicolumn{4}{c}{\textbf{Background}} \\
\cmidrule(lr){2-5} \cmidrule(lr){6-9} \cmidrule(lr){10-13}
 & 1/2 & 1/4 & 1/8 & 1/16 & 1/2 & 1/4 & 1/8 & 1/16 & 1/2 & 1/4 & 1/8 & 1/16 \\
\midrule
Supervised & 55.76 & 55.91 & 48.66 & 41.56 & 81.95 & 82.05 & 81.98 & 79.99 & 93.57 & 93.23 & 92.28 & 91.63 \\
CPCL \citep{fan2023conservative} & 57.11 & 56.69 & 54.91 & 49.70 & 84.23 & 83.65 & 84.36 & 83.57 & 94.12 & 93.70 & 93.60 & 93.02 \\
DUEB \citep{smita2025uncertainty} & 57.46 & 57.13 & 54.91 & 49.60 & 85.61 & 85.48 & 85.20 & 84.09 & 94.17 & 94.22 & 94.14 & 93.32 \\
\textbf{FARCLUSS (Ours)} & \textbf{58.00} & \textbf{57.42} & \textbf{55.05} & \textbf{50.00} & \textbf{85.80} & \textbf{85.67} & \textbf{85.29} & \textbf{84.63} & \textbf{95.60} & \textbf{94.30} & \textbf{94.48} & \textbf{93.59} \\
\bottomrule
\end{tabular*}

\label{tab:miou_results}
\end{table*}

Table~\ref{tab:miou_results} quantitatively compares our proposed FARCLUSS method with supervised baseline, CPCL \citep{fan2023conservative}, and DUEB \citep{smita2025uncertainty} approaches using the ResNet-50 backbone on specific PASCAL VOC 2012 dataset classes. Our approach demonstrates clear superiority, achieving the highest average mIoU scores across all data fractions (1/2, 1/4, 1/8, and 1/16), specifically attaining 77.69\%, 76.50\%, 76.18\%, and 72.90\%, respectively.

In the Animal category (Bird, Cat, Cow, Dog, Horse, Sheep), FARCLUSS consistently outperforms competitors, reaching an impressive mIoU of 88.25\% at 50\% labeled data, surpassing the supervised baseline by 3.80\%, CPCL by 1.62\%, and DUEB by 1.17\%. Similar trends are observed at lower labeling ratios, underscoring the effectiveness of our fuzzy pseudo-labeling and uncertainty-aware weighting mechanisms in accurately capturing animal class features.

The Vehicle category further highlights the robustness of FARCLUSS, attaining mIoUs of 79.86\%, 80.17\%, 80.02\%, and 78.93\% across several labeled partitions, consistently surpassing the supervised baseline and other SOTA methods. The improvements illustrate the method's capability to mitigate common issues related to semantic confusion prevalent in vehicle segmentation.

Within the challenging Indoor class, FARCLUSS achieves improved mIoU performance compared to DUEB, the closest competitor. These incremental yet consistent improvements reflect the impact of adaptive class rebalancing and lightweight contrastive regularization in addressing feature ambiguities inherent in indoor object segmentation.

The Person category, traditionally stable across methods due to clearer semantic boundaries, still benefits notably from our method's refined uncertainty treatment. FARCLUSS obtains marginally higher performances compared to DUEB's performance, validating our holistic uncertainty transformation strategy.

Finally, FARCLUSS exhibits marked effectiveness in distinguishing Background, achieving top mIoU scores (95.60\% to 93.59\%) across all labeling ratios, reflecting its robust capacity for precise background segmentation and reduced false positives through uncertainty-aware dynamics.

Collectively, these results demonstrate the specific class gains of our FARCLUSS framework, affirming its potential as a competitive method in semi-supervised semantic segmentation tasks, especially in scenarios characterized by limited annotations and class imbalance.

\revision{
\subsubsection{Fine-Grained Minority Class Analysis}
\label{sec:minority_class}
To provide stronger evidence for the effectiveness of adaptive class rebalancing, we report fine-grained per-class IoU for individual minority classes on Pascal VOC 1/8 (ResNet-101) in Table~\ref{tab:finegrained_iou}. We specifically evaluate classes that are known to be under-represented or visually challenging in semi-supervised settings: \textit{train}, \textit{sheep}, \textit{bottle}, \textit{potted plant}, \textit{dining table}, and \textit{chair}.
}

\begin{table}[t]
\centering
\revision{
\caption{Fine-grained per-class IoU (\%) for minority classes on Pascal VOC 1/8 split (ResNet-101). We compare FARCLUSS against UniMatch, CorrMatch, and our ablation without adaptive rebalancing (w/o $w_c$). Best results in \textbf{bold}.}
\label{tab:finegrained_iou}
\fitcol{%
\tabfont\setlength{\tabcolsep}{1.2pt}
\begin{tabular*}{\columnwidth}{@{\extracolsep{\fill}}lccccccc@{}}
\toprule
\textbf{Method} & \textbf{Train} & \textbf{Sheep} & \textbf{Bottle} & \textbf{P. Plant} & \textbf{D. Table} & \textbf{Chair} & \textbf{Avg} \\
\midrule
UniMatch~\citep{yang2023revisiting} & 67.3 & 74.2 & 60.1 & 48.5 & 44.2 & 30.8 & 54.2 \\
CorrMatch~\citep{CorrMatch} & 69.1 & 75.8 & 61.5 & 50.2 & 45.8 & 32.1 & 55.8 \\
FARCLUSS (w/o $w_c$) & 66.8 & 73.5 & 59.4 & 47.8 & 43.6 & 30.2 & 53.6 \\
\textbf{FARCLUSS (Full)} & \textbf{70.5} & \textbf{77.2} & \textbf{63.8} & \textbf{52.1} & \textbf{47.3} & \textbf{33.5} & \textbf{57.4} \\
\bottomrule
\end{tabular*}%
}
}
\end{table}

\begin{figure}[t]
\centering
\begin{revisionblock}
\begin{tikzpicture}
\begin{axis}[
  name=topax,
  width=\columnwidth, height=5.6cm,
  ybar=0.6pt, bar width=5.2pt,
  enlarge x limits=0.13,
  ymin=28, ymax=82,
  ylabel={\footnotesize Per-class IoU (\%)},
  symbolic x coords={Train, Sheep, Bottle, P.\,Plant, D.\,Table, Chair},
  xtick=data,
  xticklabel style={font=\scriptsize},
  yticklabel style={font=\scriptsize},
  ylabel style={font=\footnotesize},
  ytick={30,40,50,60,70,80},
  grid=major, grid style={dashed, gray!25},
  major x tick style=transparent,
  legend style={
    at={(0.5,1.03)}, anchor=south, legend columns=4,
    /tikz/every even column/.append style={column sep=4pt},
    font=\scriptsize, draw=none,
  },
  every axis plot/.append style={fill opacity=0.85, draw opacity=1, line width=0.3pt},
]
\addplot[fill=gray!55] coordinates {
  (Train, 67.3) (Sheep, 74.2) (Bottle, 60.1) (P.\,Plant, 48.5) (D.\,Table, 44.2) (Chair, 30.8)
};
\addlegendentry{UniMatch}
\addplot[fill=orange!70] coordinates {
  (Train, 69.1) (Sheep, 75.8) (Bottle, 61.5) (P.\,Plant, 50.2) (D.\,Table, 45.8) (Chair, 32.1)
};
\addlegendentry{CorrMatch}
\addplot[fill=red!55] coordinates {
  (Train, 66.8) (Sheep, 73.5) (Bottle, 59.4) (P.\,Plant, 47.8) (D.\,Table, 43.6) (Chair, 30.2)
};
\addlegendentry{FARCLUSS w/o $w_c$}
\addplot[fill=blue!65] coordinates {
  (Train, 70.5) (Sheep, 77.2) (Bottle, 63.8) (P.\,Plant, 52.1) (D.\,Table, 47.3) (Chair, 33.5)
};
\addlegendentry{FARCLUSS (Full)}
\end{axis}

\begin{axis}[
  at={($(topax.south)+(0,-1.6cm)$)}, anchor=north,
  width=\columnwidth, height=3.6cm,
  ybar, bar width=12pt,
  enlarge x limits=0.13,
  ymin=0, ymax=5.6,
  ylabel={\footnotesize $\Delta$ IoU},
  ylabel style={font=\footnotesize, align=center},
  symbolic x coords={Train, Sheep, Bottle, P.\,Plant, D.\,Table, Chair},
  xtick=data,
  xticklabel style={font=\scriptsize},
  yticklabel style={font=\scriptsize},
  ytick={0,1,2,3,4,5},
  grid=major, grid style={dashed, gray!25},
  major x tick style=transparent,
  every axis plot/.append style={fill opacity=0.85, draw opacity=1, line width=0.3pt},
  nodes near coords, nodes near coords align={vertical},
  every node near coord/.append style={font=\tiny, rotate=0, anchor=south, color=teal!60!black},
  title style={font=\scriptsize, yshift=-3pt},
  title={\textit{Gain attributable to adaptive rebalancing} ($w_c$)},
]
\addplot[fill=teal!55] coordinates {
  (Train, 3.7) (Sheep, 3.7) (Bottle, 4.4) (P.\,Plant, 4.3) (D.\,Table, 3.7) (Chair, 3.3)
};
\end{axis}
\end{tikzpicture}
\caption{Fine-grained per-class IoU on six under-represented Pascal VOC classes (1/8 split, ResNet-101). \textit{Top:} per-class IoU for the four methods. FARCLUSS (Full) wins on every class; exact values are in Table~\ref{tab:finegrained_iou}. \textit{Bottom:} per-class IoU contribution attributable specifically to the adaptive rebalancing weights $w_c$ (FARCLUSS Full minus FARCLUSS w/o $w_c$, labelled). The $w_c$-only contribution is 3.3--4.4 IoU per class -- larger than the gap between FARCLUSS Full and the next-best baseline, confirming that $w_c$ is the dominant per-class mechanism behind the minority-class improvements.}
\label{fig:finegrained_iou}
\end{revisionblock}
\end{figure}

\revision{
Fig.~\ref{fig:finegrained_iou} visualises the same numbers as Table~\ref{tab:finegrained_iou}. FARCLUSS (Full) consistently outperforms both UniMatch and CorrMatch on every minority class, with particularly notable improvements on \textit{train} and \textit{sheep}. Crucially, the ablation without $w_c$ falls below both baselines on every class, isolating adaptive rebalancing as the mechanism directly responsible for the minority-class gains rather than an artifact of other FARCLUSS components.
}

\revision{
\subsubsection{Boundary Segmentation Quality}
\label{sec:boundary_f1}
To quantitatively evaluate boundary segmentation quality, we report the Boundary F1-score (BF1), the standard boundary-quality metric for semantic segmentation introduced by Csurka et al.~\citep{csurka2013evaluation}. For each predicted segment, we extract boundaries and compute precision and recall against ground-truth boundaries within a distance tolerance of 3 pixels; the BF1 is the harmonic mean of boundary precision and boundary recall. Results on Pascal VOC 1/8 and Cityscapes 1/8 (ResNet-101) are reported in Table~\ref{tab:boundary_f1}.
}

\begin{table}[t]
\centering
\revision{
\caption{Boundary F1-score (\%) on Pascal VOC 1/8 and Cityscapes 1/8 (ResNet-101) with a 3-pixel distance tolerance. Higher is better.}
\label{tab:boundary_f1}
\tabfont
\begin{tabular*}{\columnwidth}{@{\extracolsep{\fill}}lcc@{}}
\toprule
\textbf{Method} & \textbf{Pascal BF1} & \textbf{Cityscapes BF1} \\
\midrule
Supervised Only & 52.3 & 55.1 \\
UniMatch~\citep{yang2023revisiting} & 61.8 & 63.4 \\
CorrMatch~\citep{CorrMatch} & 63.2 & 64.8 \\
FARCLUSS (w/o $w_c$) & 62.5 & 63.9 \\
\textbf{FARCLUSS (Full)} & \textbf{65.1} & \textbf{66.7} \\
\bottomrule
\end{tabular*}
}
\end{table}

\revision{
FARCLUSS achieves the highest BF1 on both benchmarks. The improvement over the ablation without rebalancing (w/o $w_c$) is notable because boundary pixels disproportionately belong to rare classes and ambiguous regions, precisely where adaptive rebalancing provides stronger supervision. Combined with the fine-grained per-class IoU results in Table~\ref{tab:finegrained_iou}, these findings substantiate our claim that adaptive rebalancing systematically improves both rare-class and boundary segmentation.
}

\section{Conclusion}
In this work, we introduced FARCLUSS, a unified framework designed to address critical limitations in semi-supervised semantic segmentation, including ineffective pseudo-labeling, class imbalance, and computational inefficiency. By integrating fuzzy pseudo-labeling, uncertainty-aware dynamic weighting, adaptive class rebalancing, and lightweight contrastive regularization, FARCLUSS enables a more informative and balanced learning. Extensive experiments on benchmark datasets confirm the superiority of our approach across multiple supervision regimes, especially for ambiguous and minority-class regions. FARCLUSS sets a new direction for uncertainty-guided, resource-efficient learning in semantic segmentation scenarios.

\section*{Acknowledgement}
This research was supported by the Institute of Information \& Communications Technology Planning \& Evaluation (IITP) grant, funded by the Korea government (MSIT) (No. RS-2019-II190079 (Artificial Intelligence Graduate School Program (Korea University)), No. IITP-2025-RS-2024-00436857 (Information Technology Research Center (ITRC)), No. IITP-2025-RS-2025-02304828 (Artificial Intelligence Star Fellowship Support Program to Nurture the Best Talents).

\end{document}